\documentclass[letterpaper]{article} %
\usepackage{aaai2027}  %
\usepackage[hyphens]{url}  %
\usepackage{graphicx} %
\usepackage{natbib}  %
\usepackage{caption} %
\usepackage{algorithm}
\usepackage{algorithmic}

\usepackage{newfloat}
\usepackage{listings}
\DeclareCaptionStyle{ruled}{labelfont=normalfont,labelsep=colon,strut=off} %
\floatstyle{ruled}
\newfloat{listing}{tb}{lst}{}
\floatname{listing}{Listing}

\usepackage{xcolor}
\usepackage{comment}
\usepackage{amsmath}
\usepackage{amsfonts}
\usepackage{amssymb}
\usepackage{amsthm}
\usepackage{mathtools}
\usepackage{booktabs}
\usepackage{colortbl}
\usepackage{subcaption}
\usepackage{multirow}
\usepackage{makecell}

\newtheorem{proposition}{Proposition}
\newtheorem{corollary}{Corollary}

\definecolor{rankblue}{RGB}{30, 100, 200}

\newcommand\updateremi{}
\newcommand\johanne{}
\newcommand\marianne{}

\newcommand{\vecU}{\mathbf{u}}

\newcommand{\vecY}{\mathbf{y}}

\newcommand{\CLIP}{\textrm{CLIP}}
\newcommand{\latent}{\mathbf{Z}}
\newcommand{\logit}{o}            %
\newcommand{\logits}{\mathbf{o}}  %
\newcommand{\CIout}{\mathcal{C}_{out}}
\newcommand{\nclasses}{C}             %
\newcommand{\classeName}{\cls}             %
\newcommand{\cls}{r}                  %
\newcommand{\lab}{y}                  %
\newcommand{\labs}{\mathbf{y}}        %
\newcommand{\pred}{\hat{y}}           %
\newcommand{\preds}{\hat{\mathbf{y}}} %
\newcommand{\CI}{\mathcal{C}}         %
\newcommand{\simi}{Score}                 %
\newcommand{\zsim}{z}                 %
\newcommand{\correlationmatrix}{S}   %
\newcommand{\concept}{k}
\newcommand{\conceptSet}{\mathcal{K}}
\newcommand{\loss}{\mathcal{L}oss}
\newcommand{\hh}{h}

\newcommand{\XX}{CHOQOLATE}
\newcommand{\TTI}{\textrm{CHOQ-PCR}}

\title{\XX: Organizing Concept Bottleneck Latent Spaces with Choquet Integrals}

\author{
    Rémi Kazmierczak\textsuperscript{\rm 1},
    Johanne Cohen\textsuperscript{\rm 2},
    Marianne Clausel\textsuperscript{\rm 1}
}
\affiliations{
    \textsuperscript{\rm 1}CRAN, Université de Lorraine, CNRS, Nancy, France\\
    \textsuperscript{\rm 2}LISN, Université Paris-Saclay, CNRS, Orsay, France
}

\nocopyright
\begin{document}

\maketitle
\begin{abstract}
Concept Bottleneck Models (CBMs) built on vision-language models such as
CLIP represent a latent space as human-understandable concepts. These
representations are unfaithful: related concepts are entangled, so
individual scores do not reflect their intended meaning. We propose \XX,
an interpretable-by-design layer based on 2-additive Choquet integrals,
which merges correlated concepts into compact nodes. Across four datasets,
\XX{} achieves a favorable accuracy-interpretability trade-off, with
weight-sparse and semantically coherent nodes. A closed-form gradient
derivation, backed by experiments, explains why Choquet layers drive this
organization without explicit supervision. Choquet weights also map
directly to Shapley values, which enables test-time intervention. On
standard bias-mitigation benchmarks, suppressing spurious concepts after
training performs on par with methods that require group annotations or
retraining, while needing neither.
\end{abstract}
\section{Introduction} \label{sec:intro}
\begin{figure*}[ht]
    \centering
    \begin{subfigure}[t]{0.45\linewidth}
        \centering
        \includegraphics[width=0.8\linewidth]{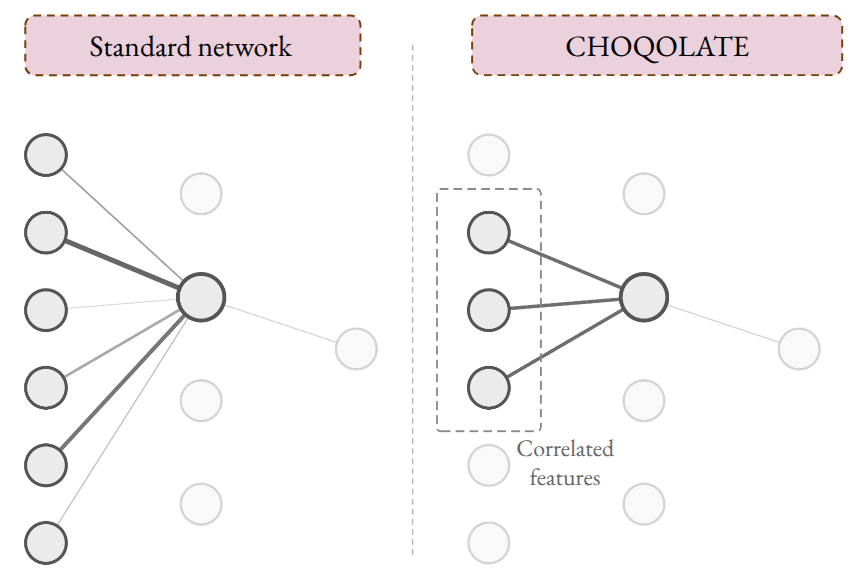}
        \vfill
        \caption{\XX{} vs.\ a standard network}
        \label{fig:teaser_a}
    \end{subfigure}  
    \begin{subfigure}[t]{0.52\linewidth}
        \centering
        \includegraphics[width=0.65\linewidth]{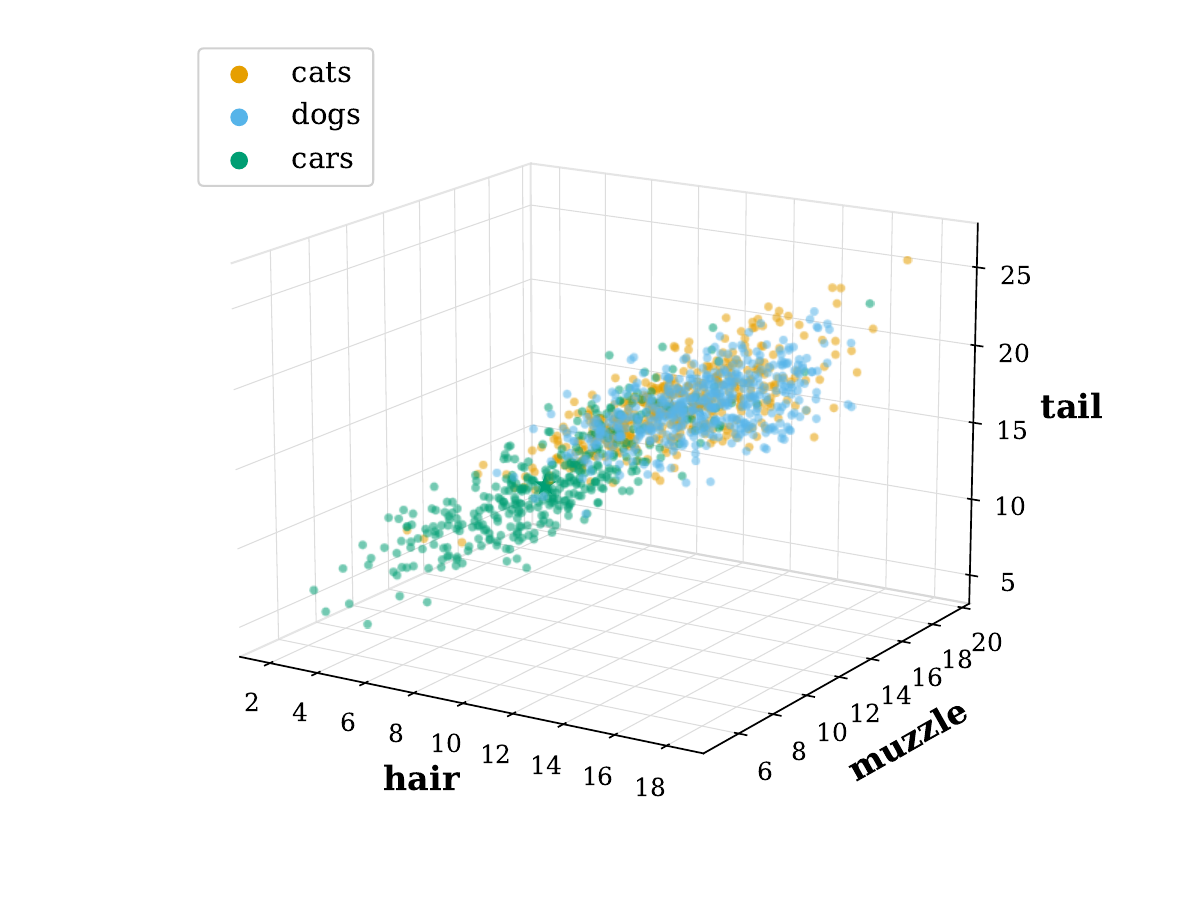}
         \caption{CLIP-similarity redundancy across related concepts}
        \label{fig:teaser_b}
    \end{subfigure}
    \caption{\textbf{Concept redundancy motivates \XX's structured design.}
\textbf{(a)} In a standard network, every input concept connects to each hidden
node, mixing concepts indiscriminately; \XX{} instead connects each node
to a small group of correlated concepts, yielding sparse and interpretable
nodes. \textbf{(b)} Visualizing a 3-concept bottleneck (\emph{hair},
\emph{muzzle}, \emph{tail}) on Cats/Dogs/Cars reveals strongly correlated
axes, with samples spreading along a shared diagonal rather than across
independent concept dimensions.}
    \label{fig:teaser}
\end{figure*}

\updateremi{Deep image classifiers reach high accuracy while offering little insight
into what drives their predictions. One influential answer routes
inference through Concept Bottleneck Models (CBMs)~\citep{koh2020concept}}, that seek
interpretability by routing inference through a layer of human-understandable concepts from which the
prediction is derived. Early
CBMs required manual concept annotations for every image. Vision-language
foundation models such as CLIP~\citep{radford2021learning} remove this
cost: the similarity between an image embedding and a textual concept
embedding directly provides a concept score, with no annotation at all~\citep{oikarinen2023labelfree,yuksekgonul2023posthoc,yang2023language}.
The annotations disappear, but a new problem takes their place: concept
unfaithfulness. Individual concept scores do not reliably reflect their
\updateremi{intended meaning~\citep{debole2025if,lewis-etal-2024-clip}}
because CLIP embeddings entangle related concepts. Figure~\ref{fig:teaser_b} illustrates the underlying redundancy on
Cats/Dogs/Cars~\citep{kazmierczak2024clip}: the CLIP similarities of \emph{hair}, \emph{muzzle}, and
\emph{tail} are strongly correlated, so a cat's \emph{tail} concept
activates even on face-only images, where no tail is visible. This mirrors the ``bag-of-words''
effect~\citep{yuksekgonul2023when}: CLIP blurs
semantically similar concepts into overlapping activations.

Rather than disentangling the backbone, we take this redundancy as
given and reorganize the concept space, fusing co-activating concepts
into a few latent nodes (Figure~\ref{fig:teaser_a}). For the resulting representation to be both
interpretable and predictive, we require five properties:
(i) \emph{attribution sparsity} (each node concentrates its attribution
on a few concepts); (ii) \emph{semantic coherence} (a node's concepts
are semantically related); (iii) \emph{attributability} (each concept's
contribution to the nodes is directly computable); (iv) \emph{low
dimensionality} (few enough nodes to compress the concept space); and
(v) \emph{informativeness} (the nodes retain what classification needs,
not everything the input contains). Properties (iii) and (iv) hold by
construction, via closed-form Shapley values and a node count far below
the concept count. Attributability, in particular, enables post-hoc
removal of unwanted concepts. Properties (i), (ii), and (v) are
quantified in Section~\ref{sec:experiments} by Attribution Gini, Node
Coherence, and accuracy, respectively.

Our tool is the 2-additive Choquet
integral~\citep{grabisch1997additive}, a classical aggregation operator
from multicriteria decision theory that combines individual weights
with pairwise interaction terms, capturing synergy when positive and
redundancy when negative. Since redundancy is precisely the structure
that CLIP similarities exhibit (Figure~\ref{fig:teaser_b}), this
operator is a natural choice for fusing co-activating concepts. We show
that layers of 2-additive Choquet integrals, trained for
classification, produce latent nodes that satisfy the five properties
above, without any supervision of the concept-to-node assignment. \XX{} stacks two such layers, one mapping concepts to a few latent nodes and one mapping nodes to class scores.

Our contributions are as follows:
\begin{itemize}
    \item We introduce \textbf{CHOQ}uet \textbf{O}rganization of
    \textbf{LATE}nts (\XX), which uses 2-additive Choquet integrals to
    build interpretable-by-design layers for concept bottleneck models.
    \item We show, across four datasets and metrics, that \XX{} offers
    a favorable trade-off between accuracy and latent-space quality.
    \item We explain why and how \XX{} organizes the latent space without
    explicit supervision, through both a closed-form gradient derivation and an empirical study of its training dynamics.
\item We show that the architecture supports \textbf{P}ost-hoc
\textbf{C}oncept \textbf{R}emoval (\TTI): on standard debiasing
benchmarks, suppressing spurious concepts after training performs on
par with methods that require group annotations or retraining, while
needing neither. 
\end{itemize}
\section{Related Work}
\subsection{Concept Bottleneck Models}
The term Concept Bottleneck Model (CBM), introduced
by~\citet{koh2020concept}, refers to the use of a bottleneck of
human-specified concepts to perform a task, predominantly image
classification, so that the resulting model is interpretable by design.
The term names a longer lineage of attribute-based
methods~\citep{kumar2009attribute,lampert2009learning}, whose early
implementations were constrained by the need for per-sample concept
annotations. Text-image contrastive foundation models have since lifted this constraint, \updateremi{enabling CBMs without explicit concept annotations~\citep{yuksekgonul2023posthoc,oikarinen2023labelfree,yang2023language}, with CLIP~\citep{radford2021learning} becoming the predominant backbone~\citep{kazmierczak2025explainability}. Several CLIP-based methods introduce hierarchical architectures, but rely on a manually specified structure~\citep{panousis2024coarse}; our approach differs in that the hierarchy emerges from training, without supervision.}
\subsection{Interpreting and Debiasing CLIP-based CBMs}
 A critical concern with foundation models is whether they base their inferences on meaningful features. This expectation is often violated, as in the ``Wolf vs.\ Husky'' example~\citep{lime}, where a classifier relies on snow rather than the object; such spurious correlations are widespread in CLIP~\citep{moayeri2024spuriosity,zhang2024common}. They directly affect CLIP-based CBMs, which suffer from \emph{concept unfaithfulness}~\citep{debole2025if}: individual concepts do not faithfully reflect what they denote, being entangled with correlated ones, and~\citet{kazmierczak2025enhancing} show that absent but semantically suggested concepts yield CLIP scores comparable to present ones. Our word-cloud representation of each latent node, i.e., a cluster of
mutually correlated concepts, makes this entanglement explicit rather
than hiding it behind a single concept score.
A parallel line of work translates embeddings into human-understandable structure, notably in mechanistic interpretability~\citep{elhage2021mathematical}. Sparse autoencoders~\citep{bricken2023monosemanticity} are popular here, with variants adding hierarchy~\citep{bussmann2025learning} or multiple latent spaces~\citep{dunefsky2024transcoders}. On the CBM side, \citet{bhalla2024interpreting} propose a task-agnostic image-embedding decomposition and \citet{rao2024discover} interpret the encoder output via a naming module; both operate on image embeddings rather than CLIP similarities,
a richer signal, but one that loses the grounding in a named concept
vocabulary on which our interpretations rely.  Among works on the CLIP-similarity space, \johanne{\citet{zhao2026partially} seek concept-class attributions, and \citet{feng2023text} target sparse attributions, but both remain
single-layer frameworks.}
Several strategies mitigate the resulting biases, most by modifying model weights. One family retrains the last layer on a group-balanced calibration set~\citep{sagawa2019distributionally,kirichenko2022last}, which is model-agnostic but requires knowing the spurious attribute per training sample. CLIP-specific methods instead leverage textual encodings:~\citet{chuang2023debiasing} project the shared embedding space to negate spurious concepts. Others~\citep{peng2026representation,debole2026concepts} modify the CLIP vision encoder directly. Our method differs from both families: removal operates post hoc on the
learned representation (\TTI), with no retraining, no per-sample group
labels, and the CLIP weights entirely unchanged.
\subsection{Choquet Integrals: From Decision Aid to Neural Networks}
Multicriteria decision aid (MCDA) studies transparent aggregation models
for high-stakes settings where a human makes the final decision; such
models are elicited from experts or learned from
data~\citep{sobrie2019learning,martyn2023deep}. Among MCDA formalisms, the Choquet integral~\citep{cho53} aggregates criteria with respect to a capacity, weighting coalitions rather than isolated criteria, and can be learned from examples~\citep{tehrani2012learning,herin2024learning}; see~\citet{grabisch2010decade} for a survey. Its 2-additive restriction~\citep{grabisch1997additive} keeps only individual weights and pairwise interactions and admits closed-form Shapley values, which model-agnostic explainers must otherwise approximate~\citep{lundberg2017unified,pelegrina2023kadditive}; these exact attributions extend to hierarchical models~\citep{labreuche2018explaining,labreuche2022explanation}. Neural formulations are recent: \citet{islam2020enabling} use fuzzy integrals as aggregation layers, and \citet{bresson2020neural} design networks whose parameter space coincides with hierarchical 2-additive Choquet integrals when the hierarchy is fixed a priori. \updateremi{More broadly, other non-linear aggregations replace the linear concept-to-logit map of standard CBMs: \citet{benard2026hoeffding} build on the Hoeffding decomposition of gradient-boosted trees for sparse, leakage-robust concept aggregation.} Closest to our work, \citet{atienza2024cutting} also map images onto CLIP-based concepts, but their Choquet model is a student distilled to explain a pixel-based teacher. \XX{} instead trains the Choquet layers as the predictor itself, with no teacher and no distillation, taking the concept space as its object of study: fusing redundant concepts into coherent nodes without supervision, explaining why this organization emerges, and supporting post-hoc concept removal (\TTI).
\section{Background} \label{sec:background}
\subsection{Concept Bottlenecks from CLIP Similarities}
\begin{figure}
    \centering
    \includegraphics[width=1\linewidth]{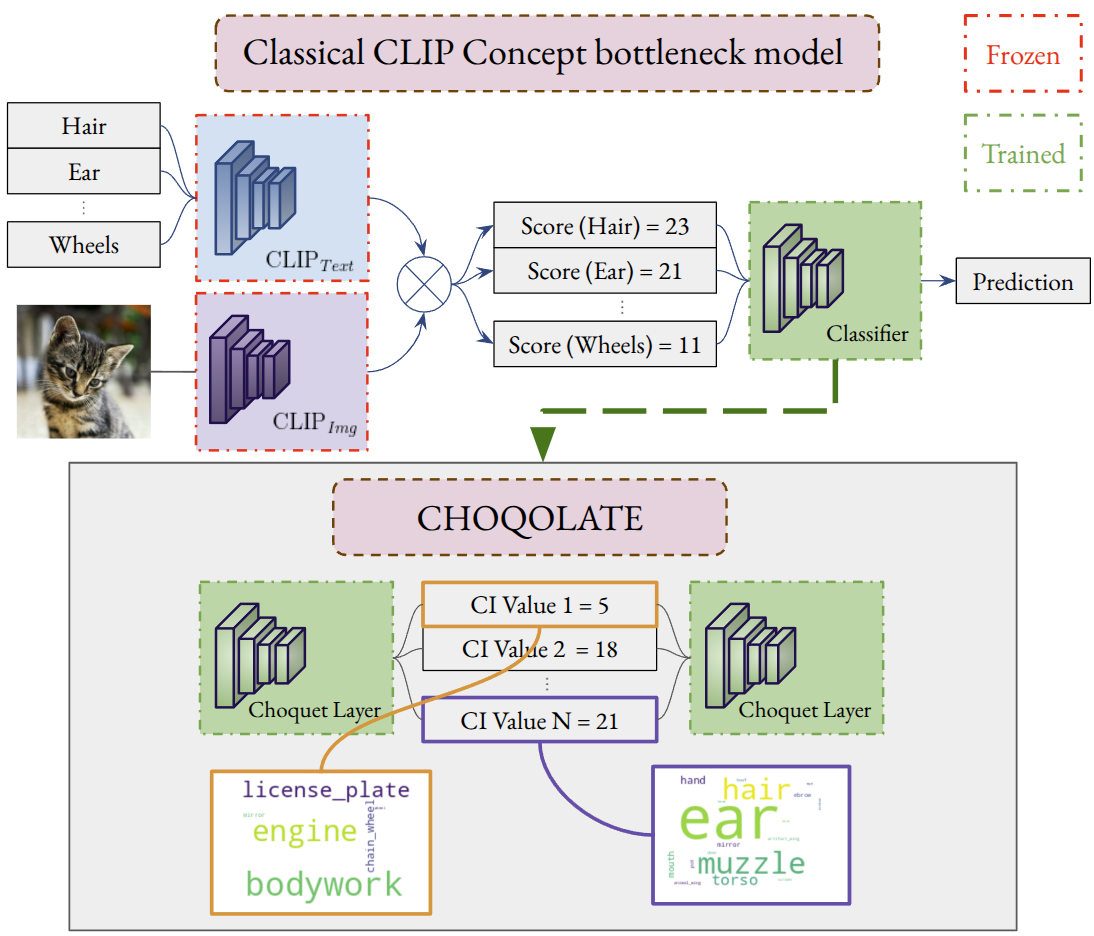}
\caption{\textbf{Principle of \XX.} Choquet layers guide the latent
space toward merging semantically similar concepts, producing sparse
and interpretable nodes, visualized as word clouds from closed-form
Shapley values.}
    \label{fig:principle_general}
\end{figure}

\paragraph{Task and notation.} We consider $\nclasses$-way image classification: an image
$I \in \mathcal{X}$ carries a one-hot label
$\labs = (\lab_1, \dots, \lab_{\nclasses}) \in \{0,1\}^{\nclasses}$.
Throughout, $j$ and $l$ index concepts (more generally, the coordinates of a
Choquet integral input), $n \in \{1, \dots, N\}$ the nodes of our
intermediate layer, and $\cls \in \{1, \dots, \nclasses\}$ the classes.
\paragraph{Concept similarities.}
A vision-language model provides two encoders into a shared $d$-dimensional
space, $\CLIP_{\text{img}} : \mathcal{X} \rightarrow \mathbb{R}^{d}$ and
$\CLIP_{\text{text}} : \mathcal{T} \rightarrow \mathbb{R}^{d}$, with
$\mathcal{T}$ a space of textual descriptions. We fix $M$ textual concepts
$\conceptSet = \{\concept_1, \dots, \concept_M\} \subset \mathcal{T}$, each
naming an attribute that may or may not be visible in an image. The \emph{CLIP similarity} between an image $I$ and $\concept_j$ is
\begin{equation}
    \simi_j \;=\; \langle \CLIP_{\text{img}}(I),\,
    \CLIP_{\text{text}}(\concept_j) \rangle.
    \label{eq:clip-score}
\end{equation}
A per-concept min-max normalization, fitted on the training set, maps this
score to $\zsim_j \in [0,1]$, which plays the role of a marginal utility in
multicriteria models~\citep{bresson2020neural}.  Stacking gives the
\emph{concept bottleneck}
$\latent = (\zsim_1, \dots, \zsim_M) \in [0,1]^{M}$, which a CLIP-based CBM
feeds to a trainable classifier, both encoders frozen. Vision-language
models entangle related concepts, so groups of coordinates co-activate;
recovering these clusters automatically is  our goal
(Figure~\ref{fig:principle_general}).
\subsection{The 2-Additive Choquet Integral}
The Choquet integral generalizes the weighted mean by letting criteria
interact. Its \emph{2-additive} form keeps only pairwise interactions,
hence a quadratic rather than exponential parameter
count~\citep{grabisch1997additive,grabisch2010decade}. We adopt the neural
formulation of~\citet{bresson2020neural}: for
$\vecU = (u_1, \dots, u_p) \in [0,1]^p$,
\begin{equation}
    \small
    \CI(\vecU) = \sum_{j=1}^{p} a_j\, u_j+ \sum_{j < l} \Bigl( b_{j,l}\, \min(u_j, u_l)+ c_{j,l}\, \max(u_j, u_l) \Bigr),
    \label{eq:choquet}
\end{equation}
where $a_j,\, b_{j,l},\, c_{j,l} \geq 0$ and
$\sum_j a_j + \sum_{j<l}(b_{j,l} + c_{j,l}) = 1$, the \emph{normalization
constraint}. These conditions make $\CI$ monotone in each coordinate,
confine its output to $[0,1]$, and cost no expressiveness: nonnegative
min/max decompositions parametrize exactly the monotone normalized
2-additive integrals~\citep{grabisch1997additive}.
The equation reads naturally: $\min(u_j, u_l)$ is a soft conjunction, large
only when both inputs are, so a positive $b_{j,l}$ encodes
\emph{complementarity}; $\max(u_j, u_l)$ is a soft disjunction, so a
positive $c_{j,l}$ encodes \emph{substitutability}~\citep{bresson2020neural};
the weights $a_j$ alone recover a weighted mean.
\paragraph{Shapley values.}
The contribution of coordinate $j$ to the output has a closed form,
classical for 2-additive measures~\citep{grabisch1997additive}:
\begin{equation}
    \mathrm{Shap}(j) \;=\; a_j + \frac{1}{2}\sum_{l \neq j}
    \bigl(b_{j,l} + c_{j,l}\bigr)\updateremi{.}
    \label{eq:shapley}
\end{equation}
A coordinate matters either on its own, through $a_j$, or through its
pairwise interactions, each counted for half.
\section{\XX} \label{sec:method}
\subsection{Architecture and Training}
\paragraph{From classifier to Choquet layers.}
\XX{} replaces the classifier of a CLIP-based CBM with two \emph{Choquet layers}: the point is not to read the concept space out but to reshape it (Figure~\ref{fig:principle_general}). The CLIP encoders stay frozen; only the Choquet weights are learned.
 \paragraph{Choquet layers.}
A \emph{Choquet layer} stacks $N$ integrals $\CI^{(1)}, \dots, \CI^{(N)}$,
each applied to the whole input vector. Each $\CI^{(n)}$ carries its own
weights $a_j^{(n)},\, b_{j,l}^{(n)},\, c_{j,l}^{(n)}$, obtained from
unconstrained parameters $\theta^{(n)}$ through a reparametrization that
enforces nonnegativity and the normalization constraint by construction
(Appendix~\ref{appendix:training}); gradient descent thus operates on
$\theta^{(n)}$ while the integral always remains a valid 2-additive
Choquet integral.
\paragraph{Two-layer architecture.}
\johanne{The \emph{first layer} takes the concept scores $\latent$ and outputs
$N \ll M$ \emph{Choquet Integral (CI) values} by grouping semantically
related concepts through learned Choquet integrals, forming an
interpretable bottleneck. The \emph{second layer} maps these $N$ CI
values to class scores via another set of Choquet integrals.
All the interpretability sits in the first layer, whose integrals we call
\emph{nodes}: the contribution of concept $\concept_j$ to node $n$ is its
Shapley value $\mathrm{Shap}^{(n)}(j)$, obtained by
applying~\eqref{eq:shapley} to the weights of $\CI^{(n)}$, rendered as
word clouds (Figure~\ref{fig:principle_general}), and edited by the
intervention of Section~\ref{subsec:tti}.}
\paragraph{Training objective.}
With $\labs$ the one-hot label and $\pred_\cls^{(T_{\mathrm{ce}})}$ the
temperature-scaled softmax of the second-layer outputs
(Appendix A), we train both layers end to end by
minimizing
\begin{equation}
    \small
    \loss^{(T_{\mathrm{ce}})} =
    \underbrace{-\sum_{\cls=1}^{\nclasses} \lab_\cls \log
        \pred_\cls^{(T_{\mathrm{ce}})}}_{\text{scaled cross-entropy}}
    +
    \lambda_{\ell_1}
    \underbrace{\sum_{n=1}^{N} \sum_{j<l}
        \bigl( b_{j,l}^{(n)} + c_{j,l}^{(n)} \bigr)}_{\ell_1 \text{ on first-layer interactions}}.
    \label{eq:loss}
\end{equation}
The $\ell_1$ penalty, of strength $\lambda_{\ell_1} \geq 0$, acts on the
first-layer interaction weights only and pushes each node toward few
pairwise interactions. 
\subsection{\TTI: Post-hoc Concept Removal} \label{subsec:tti}
Suppose a spurious attribute leaks into the prediction through a few identifiable concepts. \XX{} lets us cut them out without retraining: \TTI{} (Post-hoc Concept Removal) nullifies a concept $\concept_j$ by zeroing every first-layer weight that involves it, $a_j^{(n)} \leftarrow 0$ and $b_{j,l}^{(n)},\, c_{j,l}^{(n)} \leftarrow 0$ for all $l$ and $n$. 
\johanne{The edit is unambiguous, weights being nonnegative
and $\mathrm{Shap}^{(n)}(j)$~\eqref{eq:shapley} fully summarizing what
$\concept_j$ brings to node $n$; and it is surgical: the remaining weights
are only rescaled by a common factor to restore the normalization
constraint, so the relative contributions within each node are unchanged.}
Unlike pruning a flat linear head~\citep{yuksekgonul2023posthoc}, which removes one weight per concept, the nullification also removes every pairwise interaction of the concept, and the Shapley values quantify exactly what each node lost.
\section{Analysis of one Choquet layer} \label{sec:analysis}
We now provide insights about why Choquet layers organize the concept space
without any explicit supervision on concept clustering. To isolate the
mechanism, we study a \emph{one-layer Choquet classifier}.
Once the input $\vecU \in [0,1]^p$ is fixed, we can define
the feature vector
\begin{equation}
    \small
    \phi(\vecU) =
    \Bigl( (u_j)_{j \leq p},
           \bigl(\min(u_j, u_l)\bigr)_{j<l},
           \bigl(\max(u_j, u_l)\bigr)_{j<l} \Bigr)
    \label{eq:phi}
\end{equation}
The unnormalized weights of the Choquet layer are denoted
$\theta^{(\cls)}$; their softmax
$\mathbf{w}^{(\cls)} = \mathrm{softmax}(\theta^{(\cls)})
= (\mathbf{a}^{(\cls)}, \mathbf{b}^{(\cls)}, \mathbf{c}^{(\cls)})$
enforces the normalization constraint, and class $\cls$ is scored by a
plain inner product,
\begin{equation}
    \CI^{(\cls)}(\vecU)
    \;=\; \langle \mathbf{w}^{(\cls)}, \phi(\vecU) \rangle.
    \label{eq:onelayer}
\end{equation} 
\begin{proposition}[Choquet gradient] \label{prop:choquet_gradient}
Consider the one-layer Choquet classifier~\eqref{eq:onelayer} trained with
the \marianne{classical} cross-entropy loss $\loss$.  For every class $\cls$ and every feature
index~$i$, 
\begin{equation}
    \frac{\partial\loss}{\partial \theta_i^{(\cls)}}
    \;=\; \bigl(\hat{y}_\cls - y_\cls\bigr) \cdot w_i^{(\cls)} \cdot
    \bigl( \phi_i(\vecU) - \CI^{(\cls)}(\vecU) \bigr).
    \label{eq:deriv_result}
\end{equation}
\end{proposition}
The gradient factorizes into the classification error, the current weight
of the feature, and its deviation from the integral's output (proof in
Appendix F). Hence two mechanisms: a small
weight means a vanishing gradient, freezing unselected features out of the
dynamics; and updates concentrate on features farthest from the integral
value, inducing the anchoring of Section~\ref{sec:rq2}.
Both need a non-negligible error term, which the inner factors, of
magnitude at most one, can only shrink; hence a temperature
$T_{\mathrm{ce}} > 0$, which rescales the gradient by $1/T_{\mathrm{ce}}$
and replaces $\hat{y}_\cls$ with $\hat{y}_\cls^{(T_{\mathrm{ce}})}$
(Corollary~1,
Appendix F): a small $T_{\mathrm{ce}}$
kills the error term, and the sparsity pressure with it, once the model
is confident. In practice, class scores live in $[0,1]$, so logit gaps
never exceed one and the untempered softmax stays near uniform; our
$T_{\mathrm{ce}} = 0.005$ restores a usual logit range without going as
far as saturation.
\section{Experiments} \label{sec:experiments}
We structure our experiments around \updateremi{three} research questions:
\begin{enumerate}
    \item \textbf{RQ1:} Does \XX{} achieve a favorable trade-off between classification accuracy and latent-space interpretability?
    \item \textbf{RQ2:} Does the latent organization emerge as predicted by the gradient analysis in Section~\ref{sec:analysis}?
    \item \textbf{RQ3:} \johanne{Can post-hoc removal of spurious concepts effectively mitigate bias?} %
\end{enumerate}
\subsection{Setup}
\paragraph{Datasets.}
We evaluate on four classification datasets:
\textbf{Cats/Dogs/Cars (CDC)}~\citep{kazmierczak2024clip} (3 classes, 39
concepts), \textbf{MonumAI}~\citep{lamas2021monumai} (architectural
styles; 4 classes, 15 concepts), \textbf{COCO}~\citep{lin2014microsoft}
(location-type recognition; 6 classes, 80 concepts), and
\textbf{CUB-200-2011 (CUB)}~\citep{WahCUB_200_2011} (fine-grained bird
species; 200 classes, 226 concepts). Bias mitigation is assessed on two
binary datasets with controlled spurious correlations:
\textbf{Waterbirds}~\citep{sagawa2019distributionally} and \textbf{CelebA}~\citep{liu2015faceattributes}. Dataset constructions, spurious/non-spurious concept splits,
and full class and concept lists are given in
Appendix B
(Tables A1 and A2).
\paragraph{Metrics.}
Task performance is measured by accuracy; latent quality by two
complementary metrics, detailed in Appendix C.
\textbf{Attribution Gini}~\citep{hurley2009comparing} \marianne{taking values in $[0,1]$} quantifies the
concentration of Shapley attributions within a node: a high value
indicates a node relying on a few salient concepts, a low value a diffuse,
polysemantic one. \textbf{Node Coherence}, which we introduce, measures
whether these concepts are statistically related: %
\marianne{It is defined as the average of the pairwise correlations
$\correlationmatrix_{jl}$ between concept scores over the test set} weighted by the attributions
$\mathrm{Shap}^{(n)}(j)\cdot\mathrm{Shap}^{(n)}(l)$, so that only the
concepts a node relies on contribute. It takes values in $[-1,1]$, a high
value meaning that a node aggregates correlated, semantically related
concepts, and we report its average over the $N$ nodes. \updateremi{While not related to interpretability or classification performance, we also measure two additional
information-preservation metrics, CKNNA and HSIC, in
Appendix D.1.}
\paragraph{Baselines.}
\updateremi{As one of the most widely used competitors, we first compare against PCBM~\citep{yuksekgonul2023posthoc}. We further compare against state-of-the-art methods that emphasize learning sparse and interpretable representations, namely SLR-AVD~\citep{feng2023text}, PSCBM~\citep{zhao2026partially}, and sparse autoencoders~\citep{bricken2023monosemanticity}.}
\paragraph{Hyperparameters.}
\XX{} is trained for $100$ epochs with a batch size of $512$ and a learning rate of $0.1$. We apply an $\ell_1$ penalty with strength $\lambda_{\ell_1}=0.01$ to the first-layer interaction terms to encourage sparse concept interactions. The concept-to-class mapping uses a softmax with temperature $T_{\mathrm{ce}}=0.005$. \johanne{Unless stated otherwise, interaction terms are enabled in all reported experiments}. \updateremi{For all methods, we fix the latent dimension to $8$, chosen as a trade-off between a size large enough to preserve accuracy and one small enough to yield compact explanations; a sensitivity analysis over smaller and larger values is provided in Appendix D.2. The hyperparameter search was performed by dichotomy. The CLIP backbone is the ViT-L/14@336px model from the original CLIP release~\citep{radford2021learning}.}
\subsection{RQ1: Accuracy--Interpretability Trade-off}
\begin{figure*}[h]
    \centering
    \includegraphics[width=0.8\linewidth]{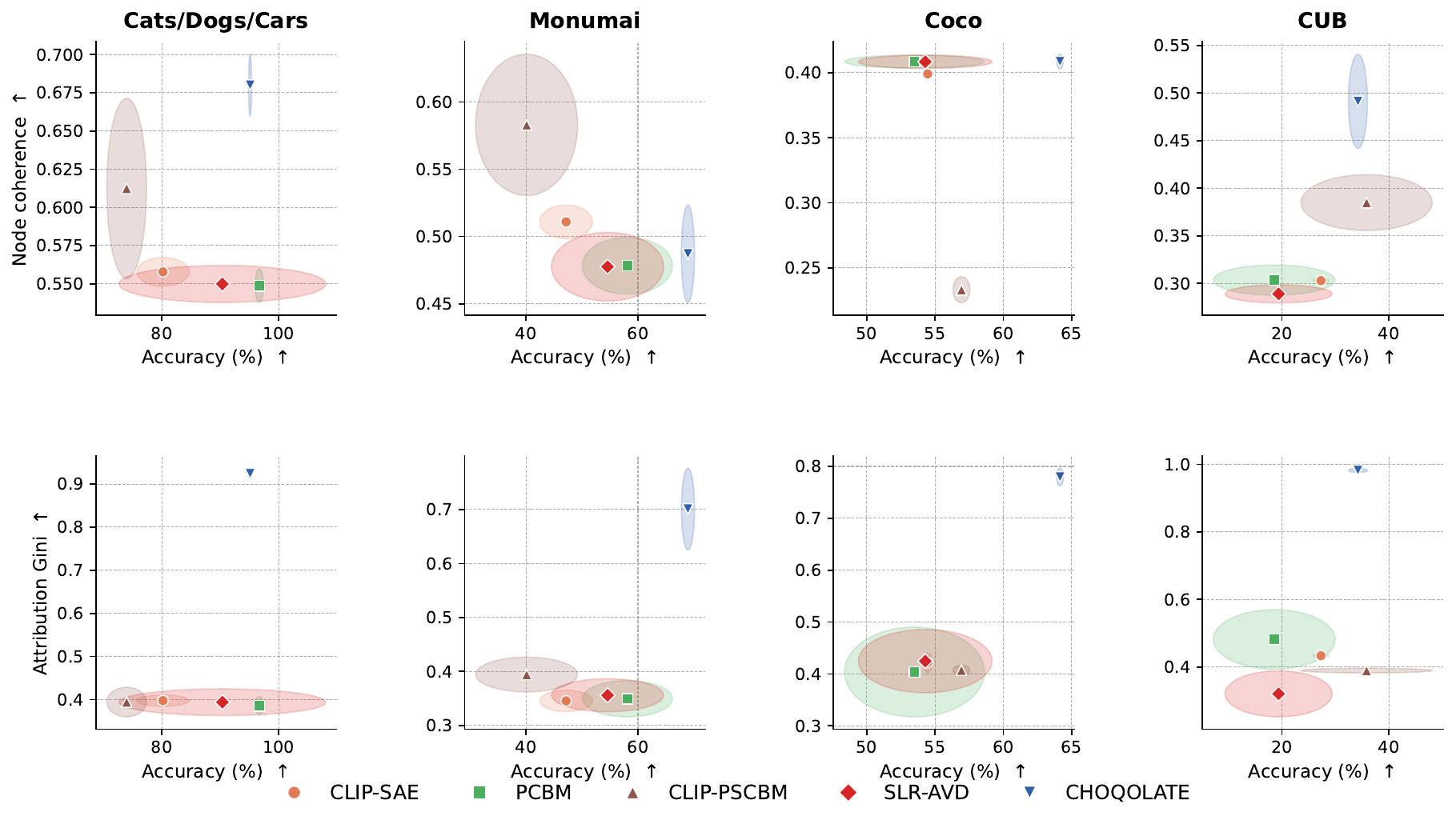}
    \caption{\textbf{Accuracy vs.\ organization-quality metrics across datasets.} Each column corresponds to one dataset (Cats/Dogs/Cars, MonumAI, COCO, CUB), each row to a quality metric. Each point is a method averaged over twenty runs, with ellipses indicating variance bounds.}
    \label{fig:results_c_vs_latent}
\end{figure*}

\begin{figure*}[ht]
    \centering
    \begin{subfigure}[t]{0.45\linewidth}
        \centering
        \includegraphics[width=\linewidth]{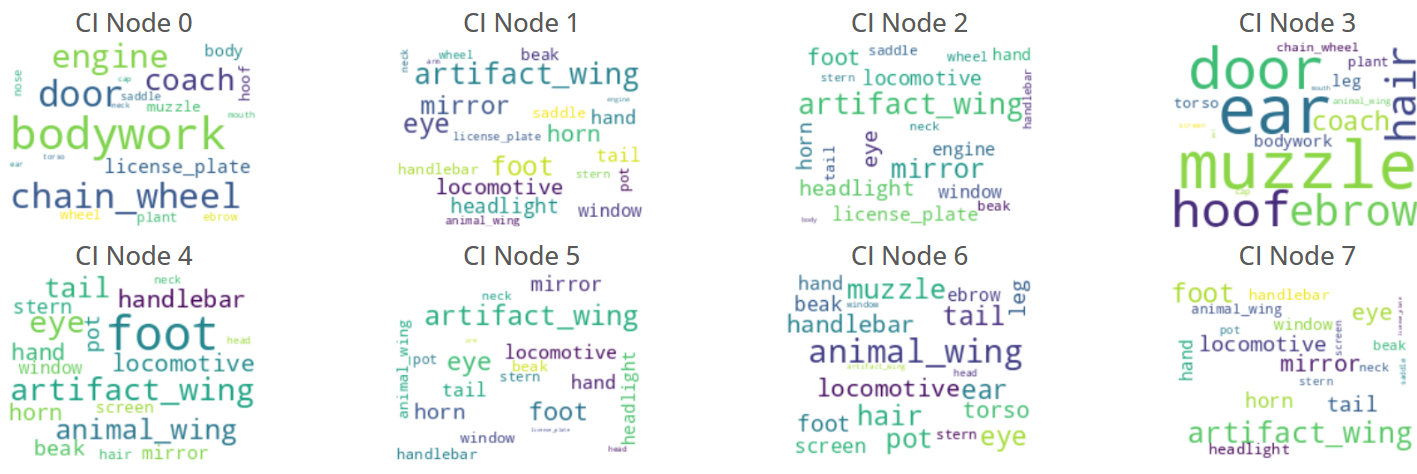}
        \caption{\updateremi{PCBM}}
        \label{fig:globalexp_mlp}
    \end{subfigure}
    \hfill
    \begin{subfigure}[t]{0.45\linewidth}
        \centering
        \includegraphics[width=\linewidth]{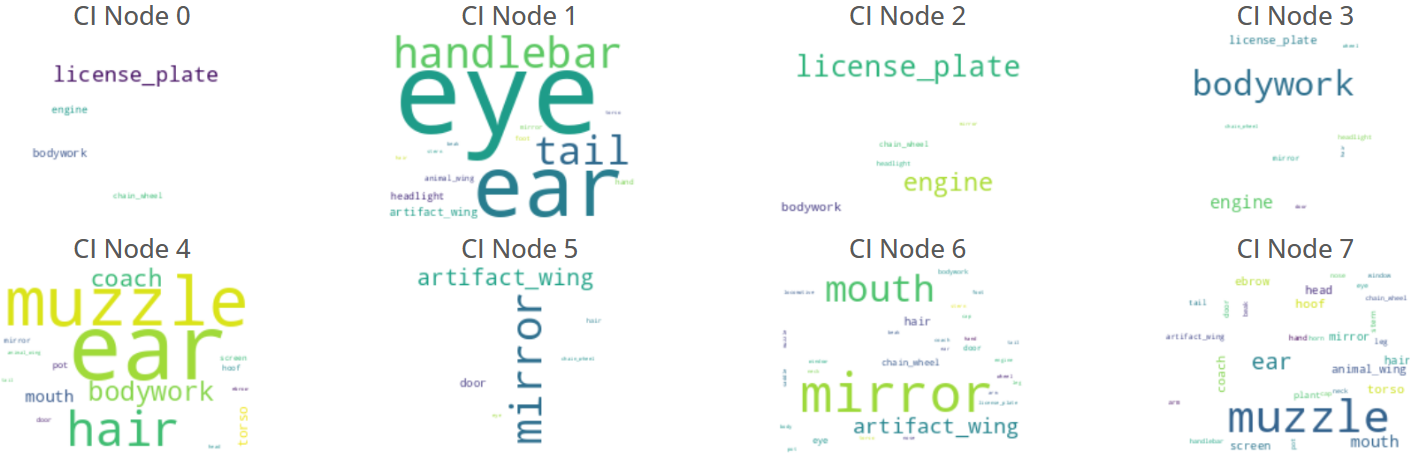}
        \caption{\XX}
        \label{fig:globalexp_choqolate}
    \end{subfigure}
\caption{\textbf{Global explanations on Cats/Dogs/Cars.} Word size
reflects concept importance for the node. \XX{} (right) yields sparser,
more coherent word clouds than the PCBM baseline (left).}
    \label{fig:globalexp}
\end{figure*}

We first ask whether \XX{} achieves a better accuracy/organization-quality
trade-off than existing interpretable representations.
Figure~\ref{fig:results_c_vs_latent} plots accuracy against each quality
metric; numerical values are in Tables~A3
and~A4 (Appendix D.1).
 To further assess the validity of these comparisons, we
report a Mann Whitney U-test in Appendix E.
On accuracy, \XX{} is on par with its competitors overall, ranking first
on MonumAI and COCO and second on Cats/Dogs/Cars and CUB
(Table A3). CUB, with 200 classes behind an 8-node
bottleneck, is hard for every method: all accuracies drop, and \XX{}
matches the best baseline ($34.20 \pm 1.56$ vs.\ $35.84 \pm 12.26$) with
a far smaller run-to-run deviation. \updateremi{We further examine how this difficulty scales with the label space in Appendix D.2.} Despite its constraints (nonnegative weights summing to one, hence monotone aggregation), \XX{} thus stays
competitive: they appear to act as a regularizing prior, and the
interaction weights add expressivity. Only on Cats/Dogs/Cars, the
simplest dataset, does this structure hinder rather than help.
Figure~\ref{fig:results_c_vs_latent} then carries our central claim:
\XX{} organizes the latent space better at comparable accuracy, sitting
in the upper-right corner of nearly every panel. It
dominates Attribution Gini everywhere, surpassing even SLR-AVD, which
applies $\ell_1$ to all weights. \updateremi{We attribute this to the
softmax reparametrization, absent from SLR-AVD, whose role is confirmed
by the ablation of Appendix D.2.} \XX{} also
achieves the best or statistically comparable Node Coherence everywhere except MonumAI, whose more
technical concept set (Table A1) is inherently
harder to cluster.
As a practical illustration, Figure~\ref{fig:globalexp} displays global
explanations as word clouds, concept size reflecting Shapley
importance~\eqref{eq:shapley}, on Cats/Dogs/Cars (more in
Appendix D.3).
The clouds of \XX{} are salient, a direct consequence of the dynamics of
Section~\ref{sec:analysis}, and semantically grouped: one node collects
\emph{ear}, \emph{muzzle}, and \emph{torso}, another \emph{bodywork},
\emph{license plate}, and \emph{engine}, shaped by the classification
task. Some redundancy across nodes indicates that the model favors few
strongly discriminative concepts over uniform coverage, suggesting
several low-utility concepts in the original vocabulary.
\subsection{RQ2: Emergence of the Organization} \label{sec:rq2}
\begin{figure*}[ht]
    \centering
    \begin{subfigure}[t]{0.4\linewidth}
        \centering
        \includegraphics[width=\linewidth]{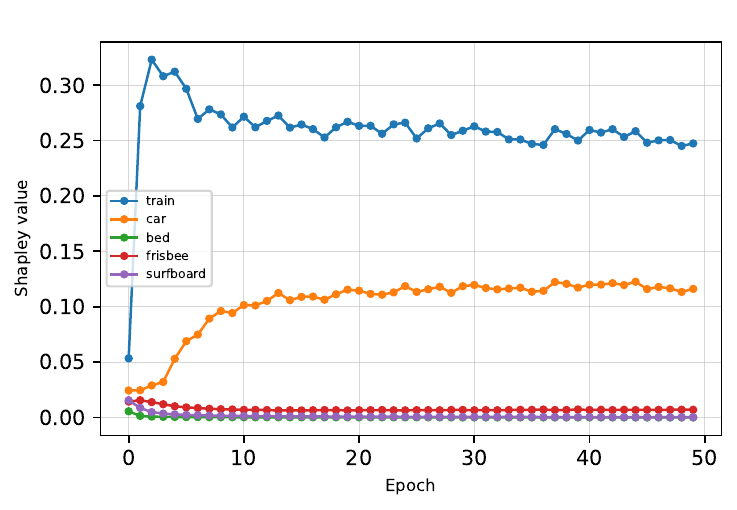}
        \caption{CI Node 2}
        \label{fig:visu_train_a}
    \end{subfigure}
    \hfill
    \begin{subfigure}[t]{0.4\linewidth}
        \centering
        \includegraphics[width=\linewidth]{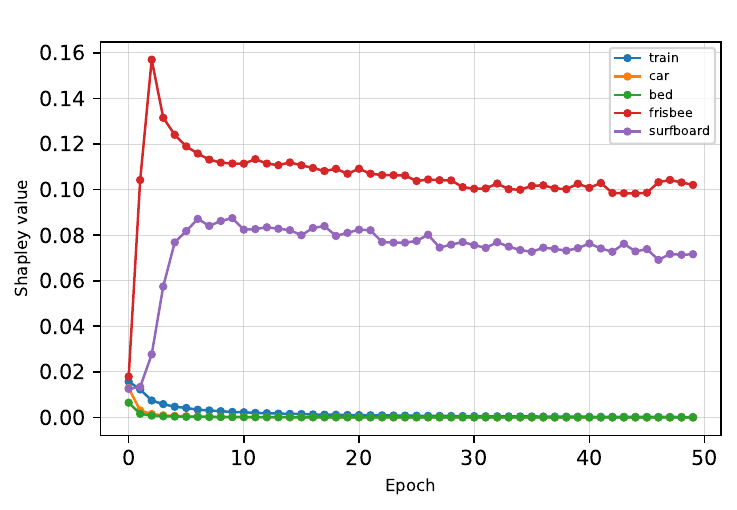}
        \caption{CI Node 4}
        \label{fig:visu_train_b}
    \end{subfigure}
    \caption{\textbf{Training dynamics of Shapley attributions on COCO.}
Per-concept attributions (Eq.~\eqref{eq:shapley}) across training epochs, in
the two nodes where the tracked concepts receive weight.}
    \label{fig:visu_train}
\end{figure*}

We next test whether the latent organization emerges as predicted by the analysis of Section~\ref{sec:analysis}. To this end, we use Equation~\eqref{eq:shapley} to track per-concept
attributions throughout training, monitoring the concepts \emph{train},
\emph{car}, \emph{bed}, \emph{frisbee}, and \emph{surfboard} on COCO,
whose concept set spans contrasted semantic groups (transportation,
leisure), and report their evolution in the nodes where they
receive non-negligible weight in Figure~\ref{fig:visu_train}.
\johanne{In both nodes, we observe a two-step process. First, a single concept,
which we refer to as the \emph{anchor} (\emph{train} in node 2 and
\emph{frisbee} in node 4), exhibits a rapid spike in attribution:
attribution mass being conserved by the softmax, an early gain comes at
the expense of all other concepts. Second, a refinement phase occurs in
which concepts semantically related to the anchor progressively gain
attribution, while unrelated concepts are driven to zero. Both steps
match Proposition~\ref{prop:choquet_gradient}: the factor 
$w_i^{(\cls)}$ freezes near-zero weights out of the dynamics, while the
deviation factor $\phi_i(\vecU) - \CI^{(\cls)}(\vecU)$ pushes down the
concepts unrelated to the anchor, whose lost weight the softmax
renormalization transfers to related ones.}

\subsection{RQ3: Post-hoc Bias Mitigation}
Finally, we ask whether the exact concept attributions of \XX{} can be
exploited after training, using bias mitigation as a case study. We apply \TTI{} (Section~\ref{subsec:tti}), assuming the spurious
concepts are known, here identified from the dataset definition, and
report accuracy and worst-group accuracy (Appendix C)
on datasets Waterbirds and CelebA, whose spurious factors are the background and
gender.

We compare  two variants of our approach. \emph{Concept removal} (oracle) knows the
spurious concepts before training and excludes them from the
bottleneck; \TTI{}  instead applies post-hoc weight nullification to a model trained on the full, biased concept set. The latter is the more practical setting, as bias is often identified only after deployment. We further compare against four established baselines: Group-DRO~\citep{sagawa2019distributionally} and DFR~\citep{kirichenko2022last},
which require group labels and fine-tuning, and the CLIP-specific
Ortho-Proj and Ortho-Cali~\citep{chuang2023debiasing}, which remove
spurious directions from the embedding space.
 Results are in Table~\ref{tab:bias_mitigation}.

\begin{table*}[ht]
    \centering
    \caption{\textbf{Accuracy (\%) of \TTI{} versus debiasing baselines.}
\textit{Baseline}: standard training. \textit{Concept removal (oracle)}:
ground-truth spurious concepts removed. \textit{\TTI{} (ours)}:
concept-weight nullification on the biased model at test time.
\textit{FT-free}: no fine-tuning; \textit{Annot.-free}: no group
annotations; \textit{Test-time}: applied at test time. Best per column
in bold.}
    \label{tab:bias_mitigation}
    \setlength{\tabcolsep}{5pt}
    \small
    \begin{tabular}{l cc cc ccc}
        \toprule
        & \multicolumn{2}{c}{\textbf{Waterbirds}}
        & \multicolumn{2}{c}{\textbf{CelebA}}
        & \multicolumn{3}{c}{\textbf{Properties}} \\
        \cmidrule(lr){2-3}\cmidrule(lr){4-5}\cmidrule(lr){6-8}
        \textbf{Method}
          & \makecell{Acc. $\uparrow$} & \makecell{Worst\\group $\uparrow$}
          & \makecell{Acc. $\uparrow$} & \makecell{Worst\\group $\uparrow$}
          & \makecell{FT\\free}
          & \makecell{Annot.\\free}
          & \makecell{Test\\time} \\
        \midrule
        Baseline
          & $84.56 \pm 1.04$ & $46.51 \pm 2.87$
          & $89.21 \pm 0.28$ & $23.89 \pm 2.72$
          & & & \\
        Concept removal (oracle)
          & $88.42 \pm 0.61$ & $60.28 \pm 3.49$
          & $91.81 \pm 0.15$ & $32.22 \pm 1.45$
          & \checkmark & \checkmark & \\
        \TTI\ (ours)
          & $\mathbf{89.52 \pm 1.12}$ & $\mathbf{68.26 \pm 10.14}$
          & $\mathbf{92.44 \pm 1.53}$ & $37.78 \pm 11.99$
          & \checkmark & \checkmark & \checkmark \\
        \midrule
        Ortho-Proj
          & $88.52 \pm 1.08$ & $58.82 \pm 3.68$
          & $91.65 \pm 0.67$ & $32.00 \pm 3.33$
          & & \checkmark & \checkmark \\
        Ortho-Cali
          & $88.61 \pm 0.33$ & $64.14 \pm 2.11$
          & $89.44 \pm 0.47$ & $11.44 \pm 4.18$
          & & \checkmark & \checkmark \\
        DFR
          & $85.65 \pm 2.02$ & $53.15 \pm 8.84$
          & $87.01 \pm 1.55$ & $37.89 \pm 2.91$
          & & & \checkmark \\
        Group-DRO
          & $82.09 \pm 2.68$ & $67.34 \pm 9.46$
          & $85.54 \pm 1.45$ & $\mathbf{47.22 \pm 3.20}$
          & & & \checkmark \\
        \bottomrule
    \end{tabular}
\end{table*}
Both settings improve substantially over the biased model, and \TTI{}
performs on par with the oracle; the two confidence intervals overlap,
so we read this as comparable rather than superior. We attribute it to two factors: training on the larger concept set may yield a richer gradient signal and better-structured representations, and nullification introduces an implicit negative signal that actively suppresses unwanted patterns rather than merely ignoring them. On worst-group accuracy, Group-DRO remains competitive and is strongest on CelebA, while our intervention is comparable on Waterbirds. The value of  \TTI{} is thus not raw dominance but its unique combination of properties: it is the only method that is simultaneously fine-tuning-free, annotation-free, and test-time. Reducing the high variance on worst-group accuracy, attributable to the small worst group, is left to future work.
\section{Limitations}
We now state the main limitations of \XX{} and, more broadly, of
Choquet layers. %
First, \XX{} operates on concept-similarity scores and relies on these scores being strongly correlated across related concepts, in line with the bag-of-words effect of vision-language models~\citep{yuksekgonul2023when,debole2025if}; this redundancy is precisely what makes merging effective. Transferring the approach to another backbone therefore requires an analogue of concept scores, and the redundancy assumption should be verified beforehand; the correlation matrix $\correlationmatrix$ used by Node Coherence provides exactly this diagnostic.
Second, even in the 2-additive case, each first-layer node carries $M + 2\binom{M}{2} = M^2$ weights, hence a quadratic dependence on the vocabulary size. This remains modest at the scale of our experiments (15 to 226 concepts) but becomes prohibitive for the tens of thousands of concepts handled by some recent CBM works~\citep{yang2023language}, and rules out settings such as SAE feature interpretation, where latent units are typically far more numerous.
Finally, since all weights are nonnegative and each integral is monotone, \XX{} constrains attributions to be nonnegative: explanations can only invoke the \emph{presence} of concepts, and evidence of absence (predicting \emph{car} because no ears are visible) never surfaces in them, even when it drives the prediction indirectly through the competition between class scores. This restriction is the price of the unambiguous edits of \TTI; the competitive accuracy of our experiments indicates that the underlying monotone aggregation suffices for the tasks we consider, and extending the framework to negative contributions is left for future work.
\section{Conclusion}
We introduced \XX, an architecture based on 2-additive Choquet integrals. Applied to CLIP-based concept bottleneck models, it offers a favorable trade-off between classification accuracy and latent-space organization. Its per-node attributions are much sparser than those of competing approaches. Our gradient analysis explains this sparsity. Under the
softmax reparametrization, the gradient of a concept is proportional to its current weight, so unselected concepts are progressively frozen out of training. This matches the anchor dynamics we observe empirically.
Choquet weights also map to Shapley values in closed form, which enables
targeted post-hoc edits: \TTI{} suppresses spurious concepts after
training and performs on par with debiasing methods that require group annotations or retraining, the edit operating directly in the interpretable concept space. 
More broadly, Choquet integrals are a promising tool for
interpretable-by-design architectures: the structure they induce emerges
without supervision, and they inherit a solid mathematical grounding from decision theory. Future work includes incorporating
Choquet layers into prototype-based models~\citep{chen2019looks} and, once
the quadratic cost in the number of inputs is alleviated, into neuron
identification methods~\citep{kalibhat2023identifying}.
\section*{Acknowledgments}
This work benefited from Hi! PARIS and State funding managed by the French National Research Agency (ANR) under the France 2030 program, reference ANR-23-IACL-0005.
This work was granted access to the HPC resources of IDRIS under the allocation AD011018094 made by GENCI.

\bibliography{aaai2027}

\begin{thebibliography}{59}
\providecommand{\natexlab}[1]{#1}

\bibitem[{Atienza et~al.(2024)Atienza, Bresson, Rousselot, Caillou, Cohen,
  Labreuche, and Sebag}]{atienza2024cutting}
Atienza, N.; Bresson, R.; Rousselot, C.; Caillou, P.; Cohen, J.; Labreuche, C.;
  and Sebag, M. 2024.
\newblock Cutting the black box: Conceptual interpretation of a deep neural net
  with multi-modal embeddings and multi-criteria decision aid.
\newblock In \emph{IJCAI 2024, 33rd International Joint Conference on
  Artificial Intelligence}, 3669--3678. International Joint Conferences on
  Artificial Intelligence Organization.

\bibitem[{B{\'e}nard et~al.(2026)B{\'e}nard, Arfib, Labreuche, and
  Qu{\'e}tu}]{benard2026hoeffding}
B{\'e}nard, C.; Arfib, M.; Labreuche, C.; and Qu{\'e}tu, V. 2026.
\newblock Hoeffding Concept Bottleneck Models with Applications to Overhead
  Images.
\newblock \emph{arXiv preprint arXiv:2606.00082}.

\bibitem[{Bhalla et~al.(2024)Bhalla, Oesterling, Srinivas, Calmon, and
  Lakkaraju}]{bhalla2024interpreting}
Bhalla, U.; Oesterling, A.; Srinivas, S.; Calmon, F.; and Lakkaraju, H. 2024.
\newblock Interpreting clip with sparse linear concept embeddings (splice).
\newblock \emph{Advances in Neural Information Processing Systems}, 37:
  84298--84328.

\bibitem[{Bresson et~al.(2020)Bresson, Cohen, H{\"u}llermeier, Labreuche, and
  Sebag}]{bresson2020neural}
Bresson, R.; Cohen, J.; H{\"u}llermeier, E.; Labreuche, C.; and Sebag, M. 2020.
\newblock Neural Representation and Learning of Hierarchical 2-additive Choquet
  Integrals.
\newblock In \emph{Proceedings of the Twenty-Ninth International Joint
  Conference on Artificial Intelligence (IJCAI)}, 1984--1991.

\bibitem[{Bricken et~al.(2023)Bricken, Templeton, Batson, Chen, Jermyn,
  Conerly, Turner, Anil, Denison, Askell, Lasenby, Wu, Kravec, Schiefer,
  Maxwell, Joseph, Hatfield-Dodds, Tamkin, Nguyen, McLean, Burke, Hume, Carter,
  Henighan, and Olah}]{bricken2023monosemanticity}
Bricken, T.; Templeton, A.; Batson, J.; Chen, B.; Jermyn, A.; Conerly, T.;
  Turner, N.~L.; Anil, C.; Denison, C.; Askell, A.; Lasenby, R.; Wu, Y.;
  Kravec, S.; Schiefer, N.; Maxwell, T.; Joseph, N.; Hatfield-Dodds, Z.;
  Tamkin, A.; Nguyen, K.; McLean, B.; Burke, J.~E.; Hume, T.; Carter, S.;
  Henighan, T.; and Olah, C. 2023.
\newblock Towards Monosemanticity: Decomposing Language Models With Dictionary
  Learning.
\newblock \emph{Transformer Circuits Thread}.

\bibitem[{Bussmann, Leask, and Nanda(2025)}]{bussmann2025learning}
Bussmann, B.; Leask, P.; and Nanda, N. 2025.
\newblock Learning Multi-Level Features with Matryoshka Sparse Autoencoders.
\newblock In \emph{International Conference on Machine Learning (ICML)}.

\bibitem[{Chen et~al.(2019)Chen, Li, Tao, Barnett, Rudin, and
  Su}]{chen2019looks}
Chen, C.; Li, O.; Tao, D.; Barnett, A.; Rudin, C.; and Su, J.~K. 2019.
\newblock This looks like that: deep learning for interpretable image
  recognition.
\newblock \emph{Advances in neural information processing systems}, 32.

\bibitem[{Choquet(1953)}]{cho53}
Choquet, G. 1953.
\newblock Theory of capacities.
\newblock \emph{Annales de l'Institut Fourier}.

\bibitem[{Chuang et~al.(2023)Chuang, Jampani, Li, Torralba, and
  Jegelka}]{chuang2023debiasing}
Chuang, C.-Y.; Jampani, V.; Li, Y.; Torralba, A.; and Jegelka, S. 2023.
\newblock Debiasing vision-language models via biased prompts.
\newblock \emph{arXiv preprint arXiv:2302.00070}.

\bibitem[{Cukierski(2013)}]{dogs-vs-cats}
Cukierski, W. 2013.
\newblock Dogs vs. Cats.

\bibitem[{Debole et~al.(2025)Debole, Barbiero, Giannini, Passerini, Teso, and
  Marconato}]{debole2025if}
Debole, N.; Barbiero, P.; Giannini, F.; Passerini, A.; Teso, S.; and Marconato,
  E. 2025.
\newblock If Concept Bottlenecks are the Question, are Foundation Models the
  Answer?
\newblock \emph{arXiv preprint arXiv:2504.19774}.

\bibitem[{Debole et~al.(2026)Debole, Passerini, Teso, Pugnana, and
  Marconato}]{debole2026concepts}
Debole, N.; Passerini, A.; Teso, S.; Pugnana, A.; and Marconato, E. 2026.
\newblock Concepts Worth Having: Refining VLM-Guided Concept Bottleneck Models
  with Minimal Annotations.
\newblock \emph{arXiv preprint arXiv:2605.16405}.

\bibitem[{Dunefsky, Chlenski, and Nanda(2024)}]{dunefsky2024transcoders}
Dunefsky, J.; Chlenski, P.; and Nanda, N. 2024.
\newblock Transcoders Find Interpretable {LLM} Feature Circuits.
\newblock In \emph{Advances in Neural Information Processing Systems
  (NeurIPS)}.

\bibitem[{Elhage et~al.(2021)Elhage, Nanda, Olsson, Henighan, Joseph, Mann,
  Askell, Bai, Chen, Conerly et~al.}]{elhage2021mathematical}
Elhage, N.; Nanda, N.; Olsson, C.; Henighan, T.; Joseph, N.; Mann, B.; Askell,
  A.; Bai, Y.; Chen, A.; Conerly, T.; et~al. 2021.
\newblock A mathematical framework for transformer circuits.
\newblock \emph{Transformer Circuits Thread}, 1(1): 12.

\bibitem[{Feng, Bair, and Kolter(2024)}]{feng2023text}
Feng, Z.; Bair, A.; and Kolter, J.~Z. 2024.
\newblock Text Descriptions are Compressive and Invariant Representations for
  Visual Learning.
\newblock \emph{Transactions on Machine Learning Research (TMLR)}.

\bibitem[{Grabisch(1997)}]{grabisch1997additive}
Grabisch, M. 1997.
\newblock K-order additive discrete fuzzy measures and their representation.
\newblock \emph{Fuzzy sets and systems}, 92(2): 167--189.

\bibitem[{Grabisch and Labreuche(2010)}]{grabisch2010decade}
Grabisch, M.; and Labreuche, C. 2010.
\newblock A decade of application of the Choquet and Sugeno integrals in
  multi-criteria decision aid.
\newblock \emph{Annals of Operations Research}, 175(1): 247--286.

\bibitem[{Gretton et~al.(2005)Gretton, Bousquet, Smola, and
  Sch{\"o}lkopf}]{gretton2005measuring}
Gretton, A.; Bousquet, O.; Smola, A.; and Sch{\"o}lkopf, B. 2005.
\newblock Measuring statistical dependence with Hilbert-Schmidt norms.
\newblock In \emph{International conference on algorithmic learning theory},
  63--77. Springer.

\bibitem[{Herin, Perny, and Sokolovska(2024)}]{herin2024learning}
Herin, M.; Perny, P.; and Sokolovska, N. 2024.
\newblock Learning preference representations based on Choquet integrals for
  multicriteria decision making.
\newblock \emph{Annals of Mathematics and Artificial Intelligence}, 92(6):
  1511--1544.

\bibitem[{Huh et~al.(2024)Huh, Cheung, Wang, and Isola}]{huh2024platonic}
Huh, M.; Cheung, B.; Wang, T.; and Isola, P. 2024.
\newblock The platonic representation hypothesis.
\newblock \emph{arXiv preprint arXiv:2405.07987}.

\bibitem[{Hurley and Rickard(2009)}]{hurley2009comparing}
Hurley, N.; and Rickard, S. 2009.
\newblock Comparing Measures of Sparsity.
\newblock \emph{IEEE Transactions on Information Theory}, 55(10): 4723--4741.

\bibitem[{Islam et~al.(2020)Islam, Anderson, Pinar, Havens, Scott, and
  Keller}]{islam2020enabling}
Islam, M.~A.; Anderson, D.~T.; Pinar, A.~J.; Havens, T.~C.; Scott, G.; and
  Keller, J.~M. 2020.
\newblock Enabling Explainable Fusion in Deep Learning with Fuzzy Integral
  Neural Networks.
\newblock \emph{IEEE Transactions on Fuzzy Systems}, 28(7): 1291--1300.

\bibitem[{Kalibhat et~al.(2023)Kalibhat, Bhardwaj, Bruss, Firooz, Sanjabi, and
  Feizi}]{kalibhat2023identifying}
Kalibhat, N.; Bhardwaj, S.; Bruss, C.~B.; Firooz, H.; Sanjabi, M.; and Feizi,
  S. 2023.
\newblock Identifying Interpretable Subspaces in Image Representations.
\newblock In \emph{International Conference on Machine Learning}, 15623--15638.

\bibitem[{Kazmierczak et~al.(2025{\natexlab{a}})Kazmierczak, Azzolin, Berthier,
  Frehse, and Franchi}]{kazmierczak2025enhancing}
Kazmierczak, R.; Azzolin, S.; Berthier, E.; Frehse, G.; and Franchi, G.
  2025{\natexlab{a}}.
\newblock Enhancing Concept Localization in CLIP-based Concept Bottleneck
  Models.
\newblock \emph{arXiv preprint arXiv:2510.07115}.

\bibitem[{Kazmierczak et~al.(2026)Kazmierczak, Azzolin, Berthier, Hedstr{\"o}m,
  Delhomme, Filliat, Bousquet, Frehse, Mancini, Caramiaux
  et~al.}]{kazmierczak2026benchmarking}
Kazmierczak, R.; Azzolin, S.; Berthier, E.; Hedstr{\"o}m, A.; Delhomme, P.;
  Filliat, D.; Bousquet, N.; Frehse, G.; Mancini, M.; Caramiaux, B.; et~al.
  2026.
\newblock Benchmarking xai explanations with human-aligned evaluations.
\newblock In \emph{Proceedings of the AAAI Conference on Artificial
  Intelligence}, volume~40, 37491--37500.

\bibitem[{Kazmierczak et~al.(2024)Kazmierczak, Berthier, Frehse, and
  Franchi}]{kazmierczak2024clip}
Kazmierczak, R.; Berthier, E.; Frehse, G.; and Franchi, G. 2024.
\newblock {CLIP-QDA}: An Explainable Concept Bottleneck Model.
\newblock \emph{Transactions on Machine Learning Research Journal}.

\bibitem[{Kazmierczak et~al.(2025{\natexlab{b}})Kazmierczak, Berthier, Frehse,
  and Franchi}]{kazmierczak2025explainability}
Kazmierczak, R.; Berthier, E.; Frehse, G.; and Franchi, G. 2025{\natexlab{b}}.
\newblock Explainability and vision foundation models: A survey.
\newblock \emph{Information Fusion}, 122: 103184.

\bibitem[{Kirichenko, Izmailov, and Wilson(2022)}]{kirichenko2022last}
Kirichenko, P.; Izmailov, P.; and Wilson, A.~G. 2022.
\newblock Last layer re-training is sufficient for robustness to spurious
  correlations.
\newblock \emph{arXiv preprint arXiv:2204.02937}.

\bibitem[{Koh et~al.(2020)Koh, Nguyen, Tang, Mussmann, Pierson, Kim, and
  Liang}]{koh2020concept}
Koh, P.~W.; Nguyen, T.; Tang, Y.~S.; Mussmann, S.; Pierson, E.; Kim, B.; and
  Liang, P. 2020.
\newblock Concept bottleneck models.
\newblock In \emph{International Conference on Machine Learning}, 5338--5348.

\bibitem[{Krause et~al.(2013)Krause, Stark, Deng, and Fei-Fei}]{krause20133d}
Krause, J.; Stark, M.; Deng, J.; and Fei-Fei, L. 2013.
\newblock 3D Object Representations for Fine-Grained Categorization.
\newblock In \emph{Proceedings of the IEEE International Conference on Computer
  Vision (ICCV) Workshops}, 554--561.

\bibitem[{Kumar et~al.(2009)Kumar, Berg, Belhumeur, and
  Nayar}]{kumar2009attribute}
Kumar, N.; Berg, A.~C.; Belhumeur, P.~N.; and Nayar, S.~K. 2009.
\newblock Attribute and simile classifiers for face verification.
\newblock In \emph{2009 IEEE 12th International Conference on Computer Vision},
  365--372. IEEE.

\bibitem[{Labreuche(2022)}]{labreuche2022explanation}
Labreuche, C. 2022.
\newblock Explanation with the Winter value: Efficient computation for
  hierarchical Choquet integrals.
\newblock \emph{International Journal of Approximate Reasoning}, 151: 225--250.

\bibitem[{Labreuche and Fossier(2018)}]{labreuche2018explaining}
Labreuche, C.; and Fossier, S. 2018.
\newblock Explaining Multi-Criteria Decision Aiding Models with an Extended
  Shapley Value.
\newblock In \emph{IJCAI}.

\bibitem[{Lamas et~al.(2021)Lamas, Tabik, Cruz, Montes, Martinez-Sevilla, Cruz,
  and Herrera}]{lamas2021monumai}
Lamas, A.; Tabik, S.; Cruz, P.; Montes, R.; Martinez-Sevilla, {\'A}.; Cruz, T.;
  and Herrera, F. 2021.
\newblock MonuMAI: Dataset, deep learning pipeline and citizen science based
  app for monumental heritage taxonomy and classification.
\newblock \emph{Neurocomputing}, 420: 266--280.

\bibitem[{Lampert, Nickisch, and Harmeling(2009)}]{lampert2009learning}
Lampert, C.~H.; Nickisch, H.; and Harmeling, S. 2009.
\newblock Learning to detect unseen object classes by between-class attribute
  transfer.
\newblock In \emph{2009 IEEE Conference on Computer Vision and Pattern
  Recognition}, 951--958. IEEE.

\bibitem[{Lewis et~al.(2024)Lewis, Nayak, Yu, Merullo, Yu, Bach, and
  Pavlick}]{lewis-etal-2024-clip}
Lewis, M.; Nayak, N.; Yu, P.; Merullo, J.; Yu, Q.; Bach, S.; and Pavlick, E.
  2024.
\newblock Does {CLIP} Bind Concepts? Probing Compositionality in Large Image
  Models.
\newblock In Graham, Y.; and Purver, M., eds., \emph{Findings of the
  Association for Computational Linguistics: EACL 2024}, 1487--1500. St.
  Julian{'}s, Malta: Association for Computational Linguistics.

\bibitem[{Lin et~al.(2014)Lin, Maire, Belongie, Hays, Perona, Ramanan,
  Doll{\'a}r, and Zitnick}]{lin2014microsoft}
Lin, T.-Y.; Maire, M.; Belongie, S.; Hays, J.; Perona, P.; Ramanan, D.;
  Doll{\'a}r, P.; and Zitnick, C.~L. 2014.
\newblock Microsoft {COCO}: Common Objects in Context.
\newblock In \emph{European Conference on Computer Vision (ECCV)}, 740--755.
  Springer.

\bibitem[{Liu et~al.(2015)Liu, Luo, Wang, and Tang}]{liu2015faceattributes}
Liu, Z.; Luo, P.; Wang, X.; and Tang, X. 2015.
\newblock Deep Learning Face Attributes in the Wild.
\newblock In \emph{Proceedings of International Conference on Computer Vision
  (ICCV)}.

\bibitem[{Lundberg and Lee(2017)}]{lundberg2017unified}
Lundberg, S.~M.; and Lee, S.-I. 2017.
\newblock A unified approach to interpreting model predictions.
\newblock \emph{Advances in neural information processing systems}, 30.

\bibitem[{Martyn and Kadzi{\'n}ski(2023)}]{martyn2023deep}
Martyn, K.; and Kadzi{\'n}ski, M. 2023.
\newblock Deep preference learning for multiple criteria decision analysis.
\newblock \emph{European Journal of Operational Research}, 305(2): 781--805.

\bibitem[{Moayeri et~al.(2023)Moayeri, Wang, Singla, and
  Feizi}]{moayeri2024spuriosity}
Moayeri, M.; Wang, W.; Singla, S.; and Feizi, S. 2023.
\newblock Spuriosity rankings: Sorting data to measure and mitigate biases.
\newblock \emph{Advances in Neural Information Processing Systems}, 36:
  41572--41600.

\bibitem[{Oikarinen et~al.(2023)Oikarinen, Das, Nguyen, and
  Weng}]{oikarinen2023labelfree}
Oikarinen, T.; Das, S.; Nguyen, L.~M.; and Weng, T.-W. 2023.
\newblock Label-free Concept Bottleneck Models.
\newblock In \emph{The Eleventh International Conference on Learning
  Representations, {ICLR} 2023}.

\bibitem[{Panousis, Ienco, and Marcos(2024)}]{panousis2024coarse}
Panousis, K.~P.; Ienco, D.; and Marcos, D. 2024.
\newblock Coarse-to-fine concept bottleneck models.
\newblock \emph{Advances in Neural Information Processing Systems}, 37:
  105171--105199.

\bibitem[{Pelegrina, Duarte, and Grabisch(2023)}]{pelegrina2023kadditive}
Pelegrina, G.~D.; Duarte, L.~T.; and Grabisch, M. 2023.
\newblock A $k$-additive {Choquet} Integral-Based Approach to Approximate the
  {SHAP} Values for Local Interpretability in Machine Learning.
\newblock \emph{Artificial Intelligence}, 325: 104014.

\bibitem[{Peng et~al.(2026)Peng, Xie, Hao, Jin, and
  Huang}]{peng2026representation}
Peng, P.; Xie, M.-K.; Hao, H.; Jin, T.; and Huang, S.-J. 2026.
\newblock Representation-Level Counterfactual Calibration for Debiased
  Zero-Shot Recognition.
\newblock \emph{Advances in Neural Information Processing Systems}, 38:
  134547--134584.

\bibitem[{Radford et~al.(2021)Radford, Kim, Hallacy, Ramesh, Goh, Agarwal,
  Sastry, Askell, Mishkin, Clark et~al.}]{radford2021learning}
Radford, A.; Kim, J.~W.; Hallacy, C.; Ramesh, A.; Goh, G.; Agarwal, S.; Sastry,
  G.; Askell, A.; Mishkin, P.; Clark, J.; et~al. 2021.
\newblock Learning transferable visual models from natural language
  supervision.
\newblock In \emph{International conference on machine learning}, 8748--8763.
  PmLR.

\bibitem[{Rao(1982)}]{rao1982diversity}
Rao, C.~R. 1982.
\newblock Diversity and dissimilarity coefficients: a unified approach.
\newblock \emph{Theoretical population biology}, 21(1): 24--43.

\bibitem[{Rao et~al.(2024)Rao, Mahajan, B{\"o}hle, and
  Schiele}]{rao2024discover}
Rao, S.; Mahajan, S.; B{\"o}hle, M.; and Schiele, B. 2024.
\newblock Discover-then-Name: Task-Agnostic Concept Bottlenecks via Automated
  Concept Discovery.
\newblock In \emph{European Conference on Computer Vision (ECCV)}.

\bibitem[{Ribeiro, Singh, and Guestrin(2016)}]{lime}
Ribeiro, M.~T.; Singh, S.; and Guestrin, C. 2016.
\newblock ``{W}hy Should {I} Trust You?'': Explaining the Predictions of Any
  Classifier.
\newblock In \emph{Proceedings of the 22nd {ACM} {SIGKDD} International
  Conference on Knowledge Discovery and Data Mining, San Francisco, CA, USA,
  August 13-17, 2016}, 1135--1144.

\bibitem[{Sagawa et~al.(2019)Sagawa, Koh, Hashimoto, and
  Liang}]{sagawa2019distributionally}
Sagawa, S.; Koh, P.~W.; Hashimoto, T.~B.; and Liang, P. 2019.
\newblock Distributionally robust neural networks for group shifts: On the
  importance of regularization for worst-case generalization.
\newblock \emph{arXiv preprint arXiv:1911.08731}.

\bibitem[{Sobrie, Mousseau, and Pirlot(2019)}]{sobrie2019learning}
Sobrie, O.; Mousseau, V.; and Pirlot, M. 2019.
\newblock Learning monotone preferences using a majority rule sorting model.
\newblock \emph{International Transactions in Operational Research}, 26(5):
  1786--1809.

\bibitem[{Tehrani et~al.(2012)Tehrani, Cheng, Dembczy\'{n}ski, and
  H\"{u}llermeier}]{tehrani2012learning}
Tehrani, A.~F.; Cheng, W.; Dembczy\'{n}ski, K.; and H\"{u}llermeier, E. 2012.
\newblock Learning monotone nonlinear models using the Choquet integral.
\newblock \emph{Machine learning}.

\bibitem[{Wah et~al.(2011)Wah, Branson, Welinder, Perona, and
  Belongie}]{WahCUB_200_2011}
Wah, C.; Branson, S.; Welinder, P.; Perona, P.; and Belongie, S. 2011.
\newblock Caltech-UCSD Birds-200-2011.
\newblock Technical Report CNS-TR-2011-001, California Institute of Technology.

\bibitem[{Yang et~al.(2023)Yang, Panagopoulou, Zhou, Jin, Callison-Burch, and
  Yatskar}]{yang2023language}
Yang, Y.; Panagopoulou, A.; Zhou, S.; Jin, D.; Callison-Burch, C.; and Yatskar,
  M. 2023.
\newblock Language in a bottle: Language model guided concept bottlenecks for
  interpretable image classification.
\newblock In \emph{Proceedings of the IEEE/CVF conference on computer vision
  and pattern recognition}, 19187--19197.

\bibitem[{Yuksekgonul et~al.(2023)Yuksekgonul, Bianchi, Kalluri, Jurafsky, and
  Zou}]{yuksekgonul2023when}
Yuksekgonul, M.; Bianchi, F.; Kalluri, P.; Jurafsky, D.; and Zou, J. 2023.
\newblock When and why vision-language models behave like bags-of-words, and
  what to do about it?
\newblock In \emph{International Conference on Learning Representations
  (ICLR)}.

\bibitem[{Yuksekgonul, Wang, and Zou(2023)}]{yuksekgonul2023posthoc}
Yuksekgonul, M.; Wang, M.; and Zou, J. 2023.
\newblock Post-hoc Concept Bottleneck Models.
\newblock In \emph{International Conference on Learning Representations
  (ICLR)}.

\bibitem[{Zhang et~al.(2024)Zhang, Colman, Shahriyari, Bharaj
  et~al.}]{zhang2024common}
Zhang, M.; Colman, B.; Shahriyari, A.; Bharaj, G.; et~al. 2024.
\newblock Common-Sense Bias Discovery and Mitigation for Classification Tasks.
\newblock \emph{arXiv preprint arXiv:2401.13213}.

\bibitem[{Zhao et~al.(2026)Zhao, Huang, Yan, Sun, and Yu}]{zhao2026partially}
Zhao, D.; Huang, Q.; Yan, D.; Sun, Y.; and Yu, J. 2026.
\newblock Partially shared concept bottleneck models.
\newblock In \emph{Proceedings of the AAAI Conference on Artificial
  Intelligence}, volume~40, 13117--13125.

\bibitem[{Zhou et~al.(2016)Zhou, Khosla, Lapedriza, Torralba, and
  Oliva}]{zhou2016places}
Zhou, B.; Khosla, A.; Lapedriza, A.; Torralba, A.; and Oliva, A. 2016.
\newblock Places: An image database for deep scene understanding.
\newblock \emph{arXiv preprint arXiv:1610.02055}.

\end{thebibliography}

\clearpage

\appendix
\renewcommand{\thetable}{A\arabic{table}}
\renewcommand{\thefigure}{A\arabic{figure}}
\renewcommand{\theequation}{A\arabic{equation}}
\renewcommand{\thealgorithm}{A\arabic{algorithm}}
\setcounter{table}{0}
\setcounter{figure}{0}
\setcounter{equation}{0}

\onecolumn

\section{\updateremi{Weight Parametrization and Training Procedure}}
\label{appendix:training}
\label{appendix:softmax} %
Each Choquet integral is trained without any explicit constraint: its
weights are the softmax of an unconstrained parameter vector
$\theta \in \mathbb{R}^{q}$, with $q = p + 2\binom{p}{2} = p^2$,
\begin{equation}
    \mathbf{w} \;=\; (\mathbf{a}, \mathbf{b}, \mathbf{c}) \;=\; \mathrm{softmax}(\theta),
    \label{eq:softmax_reparam}
\end{equation}
which reads coordinatewise as
\begin{equation}
    a_j \;=\; \frac{\exp(\theta^a_j)}{\Omega}, \qquad
    b_{j,l} \;=\; \frac{\exp(\theta^b_{j,l})}{\Omega}, \qquad
    c_{j,l} \;=\; \frac{\exp(\theta^c_{j,l})}{\Omega},
    \label{eq:softmax_reparam_appendix}
\end{equation}
where $\Omega = \sum_{j=1}^{p} \exp(\theta^a_j)
+ \sum_{j<l} \bigl(\exp(\theta^b_{j,l}) + \exp(\theta^c_{j,l})\bigr)$ is the
common normalizer. Nonnegativity and the sum-to-one condition hold by
construction, so the normalization constraint never needs to be projected or
penalized: training is plain gradient descent on the parameters.

\paragraph{\XX{} architecture.}
\XX{} chains two Choquet layers on top of the frozen CLIP encoders
$\CLIP_{\text{img}}$ and $\CLIP_{\text{text}}$, which only serve to produce
the concept vector $\latent$ (Algorithm~\ref{alg:training}, line~1).
 The first
maps the concept vector $\latent$ to the CI values
$\mathbf{h} = (h_1, \dots, h_N)$, with $h_n = \CI^{(n)}(\latent)$; the
second maps $\mathbf{h}$ to the logit vector
$\logits = (\logit_1, \dots, \logit_{\nclasses})$,
\begin{equation}
    \logit_\cls \;=\; \CIout^{(\cls)}(\mathbf{h}),
    \qquad
    \cls = 1, \dots, \nclasses,
    \label{eq:logits}
\end{equation}
and a temperature-scaled softmax turns the logits into predicted
probabilities,
$\pred_\cls^{(T_{\mathrm{ce}})}
= \mathrm{softmax}(\logits / T_{\mathrm{ce}})_\cls$ with
$T_{\mathrm{ce}} > 0$. We train both layers end to end, encoders frozen, by
minimizing
\begin{equation*}
    \loss \;=\;
    \underbrace{-\sum_{\cls=1}^{\nclasses} \lab_\cls \log
        \pred_\cls^{(T_{\mathrm{ce}})}}_{\text{cross-entropy}}
    \;+\;
    \lambda_{\ell_1}
    \underbrace{\sum_{n=1}^{N} \sum_{j<l}
        \bigl( b_{j,l}^{(n)} + c_{j,l}^{(n)} \bigr)}_{\ell_1 \text{ on first-layer interactions}},
    \hspace{9cm} (4)
\end{equation*}
where $\labs$ is the one-hot label. The cross-entropy term fits the labels;
the $\ell_1$ penalty, of strength $\lambda_{\ell_1} \geq 0$, acts on the
first-layer interaction weights only, pushing each node toward few pairwise
interactions.

Algorithm~\ref{alg:training} summarizes the procedure. The encoders being
frozen, the concept similarities are computed once before training; each
step then updates only the parameters of the $N + \nclasses$ integrals,
that is, $NM^2 + \nclasses N^2$ scalars in total.

\begin{algorithm}
\caption{Training procedure of CHOQOLATE}
\label{alg:training}
\begin{algorithmic}[1]
\REQUIRE training images with one-hot labels $\labs$; concepts $\conceptSet$; frozen encoders $\CLIP_{\text{img}}, \CLIP_{\text{text}}$; nodes $N$; temperature $T_{\mathrm{ce}}$; penalty $\lambda_{\ell_1}$
\STATE \textbf{Precompute} for every image: similarities $\simi_j$~(1), then $\zsim_j$ by per-concept min-max normalization fit on the training set; store $\latent = (\zsim_1, \dots, \zsim_M)$
\STATE \textbf{Initialize} the unconstrained parameters $\theta$ of the $N + \nclasses$ integrals
\FOR{each epoch}
\FOR{each mini-batch}
\STATE $\mathbf{w} = (a, b, c) \gets \mathrm{softmax}(\theta)$ for every integral \COMMENT{Eq.~\eqref{eq:softmax_reparam}}
\STATE $h_n \gets \CI^{(n)}(\latent)$ for $n = 1, \dots, N$ \COMMENT{CI values}
\STATE $\logit_\cls \gets \CIout^{(\cls)}(\mathbf{h})$ for $\cls = 1, \dots, \nclasses$ \COMMENT{Eq.~\eqref{eq:logits}}
\STATE $\preds^{(T_{\mathrm{ce}})} \gets \mathrm{softmax}(\logits / T_{\mathrm{ce}})$
\STATE $\loss \gets -\sum_{\cls} \lab_\cls \log \pred_\cls^{(T_{\mathrm{ce}})} + \lambda_{\ell_1} \sum_{n}\sum_{j<l} \bigl(b^{(n)}_{j,l} + c^{(n)}_{j,l}\bigr)$ \COMMENT{Eq.~(4)}
\STATE one gradient step on all unconstrained parameters $\theta$
\ENDFOR
\ENDFOR
\RETURN the two trained layers, and the attributions $\mathrm{Shap}^{(n)}(j)$~(3) for interpretation
\end{algorithmic}
\end{algorithm}

\section{Dataset Details} \label{appendix:datasets}
We evaluate on four classification datasets, listed by increasing number of classes:
\begin{itemize}
    \item  \textbf{Cats/Dogs/Cars (CDC)}~\citep{kazmierczak2024clip}, built from the Kaggle Cats and Dogs~\citep{dogs-vs-cats} and Stanford Cars~\citep{krause20133d} datasets (3 classes, 39 concepts);  \item\textbf{MonumAI}~\citep{lamas2021monumai}, architectural style classification from facade photographs (4 classes, 15 concepts);  \item\textbf{COCO}~\citep{lin2014microsoft}, where the task is location-type recognition in everyday scenes (6 classes, the 80 standard COCO object categories as concepts);
    \item  \textbf{CUB-200-2011 (CUB)}~\citep{WahCUB_200_2011}, fine-grained bird classification widely used in the CBM literature (200 classes, 226 concepts). 
    \end{itemize} 
Bias mitigation is assessed on two binary datasets with controlled spurious correlations:
\begin{itemize}
    \item \textbf{Waterbirds}~\citep{sagawa2019distributionally} composites CUB birds onto Places backgrounds~\citep{zhou2016places}, with a strong (95\%) correlation between bird type and background at training time and a balanced test set (48 non-spurious, 9 spurious concepts).
    \item \textbf{CelebA}~\citep{liu2015faceattributes} targets the hair-color/gender correlation, with identical spurious-feature proportions across splits (35 non-spurious, 12 spurious concepts).
\end{itemize} 
    
\updateremi{Tables~\ref{tab:dataset_details} and~\ref{tab:spurious_dataset_details} list the classes and human-annotated concepts associated with each dataset used in our experiments. 
  Concepts were selected following the procedure of~\cite{kazmierczak2026benchmarking} and correspond to visually grounded attributes physically present in the image, so as to minimize the ambiguity inherent in more abstract descriptions.}

\begin{table*}[h]
    \centering
    \small
    \renewcommand{\arraystretch}{1.3}
    \updateremi{\begin{tabular}{lp{0.15\linewidth}p{0.7\linewidth}}
        \toprule
        \textbf{Dataset} & \textbf{Classes} & \textbf{Concepts} \\
        \midrule
        Cats-Dogs-Cars 
        & cats, dogs, cars 
        & engine, artifact wing, animal wing, stern, tail, locomotive, arm, hair, wheel, chain wheel, handlebar, hand, headlight, saddle, body, bodywork, beak, head, eye, foot, leg, neck, torso, cap, license plate, door, mirror, window, ear, muzzle, horn, nose, hoof, mouth, eyebrow, plant, pot, coach, screen \\
        \midrule
        MonumAI 
        & Baroque, Gothic, Hispanic-Muslim, Renaissance 
        & horseshoe arch, lobed arch, pointed arch, ogee arch, trefoil arch, serliana, solomonic column, pinnacle gothic, porthole, broken pediment, rounded arch, flat arch, segmental pediment, triangular pediment, lintelled doorway \\
        \midrule
        COCO 
        & shopping and dining, workplace, home or hotel, transportation, sports and leisure, cultural 
        & person, backpack, umbrella, handbag, tie, suitcase, bicycle, car, motorcycle, airplane, bus, train, truck, boat, traffic light, fire hydrant, stop sign, parking meter, bench, bird, cat, dog, horse, sheep, cow, elephant, bear, zebra, giraffe, frisbee, skis, snowboard, sports ball, kite, baseball bat, baseball glove, skateboard, surfboard, tennis racket, bottle, wine glass, cup, fork, knife, spoon, bowl, banana, apple, sandwich, orange, broccoli, carrot, hot dog, pizza, donut, cake, chair, couch, potted plant, bed, dining table, toilet, tv, laptop, mouse, remote, keyboard, cell phone, microwave, oven, toaster, sink, refrigerator, book, clock, vase, scissors, teddy bear, hair drier, toothbrush \\
        \midrule
        CUB-200-2011 
        & 200 fine-grained bird species (e.g., Black-footed Albatross, Indigo Bunting, Brown Pelican, Winter Wren) ; full list at~\url{https://github.com/yossigandelsman/clip_text_span/blob/main/utils/cub_classes.py} 
        & a short, conical bill; a large, conical bill; a stout, conical bill; a long, thin bill; a long, curved bill; a thin, curved bill; a hooked bill; a long, hooked bill; a short, hooked beak; a short, stubby bill; a short, blunt beak; a long, pointed bill; a long, straight beak; a long, thick bill; a large bill; a small bill; a black bill; a yellow bill; a pink bill; a large, orange bill; a dark, glossy bill; a yellow bill with a red spot; a red spot on the beak; a black cap; a black hood; a black cap and bib; a brown cap; a rusty-brown cap; a red cap on the head; a gray crown; a yellow crown; a white crown; a rust-colored cap and nape; a brown cap and white eyebrows; a crest on the head; a black crest on the head; a red crest on the head; a pointy crest on the head; a black mask over the eyes; a black mask across the face; a black eye stripe; a black line through the eye; a black stripe on the head; a pale stripe above the eye; a white eyebrow; a white stripe over the eye; a black head; a black head and neck; a black face; a gray head; a red head; a blue head; a green head; a yellow head; a black and white striped head; a violet ear patch; all black plumage; all-white plumage; a black and white plumage; a gray and white plumage; a brown and white plumage; a black and orange plumage; spotted plumage; a streaked brown plumage; a brownish-gray plumage; grayish-brown plumage; a brown body; a gray body; a gray plumage; a dark coloration; a reddish-brown body; a grayish-brown body; a grayish-white body; a blue-gray body; a bright yellow body; a bright green body; a green body; iridescent plumage; iridescent green plumage; iridescent black feathers; iridescent blue-green back; a glossy, black plumage; bright golden-yellow plumage; a vibrant blue plumage; a Scarlet-red body; a bright red breast; a rusty-red breast; a bright orange breast; a rosy breast; a pinkish breast; a buffy breast; a white breast; a gray breast; a brown breast; a yellow breast; a black breast; a streaked breast; a black band across the breast; a black crescent on the breast; dark streaks on the breast; a ruby-red throat; a red throat; a yellow throat; a white throat; a white chin and throat; a yellow throat and breast; a orange breast and belly; a white belly; a yellow belly; a pale belly; a red belly; white underparts; grayish-white underparts; a white underbelly; a light-colored belly; a white rump; a white patch on the wing; a brown back; a black back; a gray back; a grayish-brown back; a reddish-brown back; a brownish back; a dark gray back; a green back; a blue back; iridescent blue-green back; blue upperparts; brown upperparts; gray upperparts; dark blue-black upperparts; olive-green upperparts; pale blue-gray upperparts; greenish upperparts; olive-gray upperparts; a streaked back; a gray back with black streaks; a brown back with dark streaks; black wingtips; dark wingtips; white wing bars; two white bars on the wings; black markings on the wings; white patches on the wings; white stripes on the wings; black wings with white bars; brown wings with white bars; black wings with orange bars; orange and black wings; blue wings with black bars; blue wings; brown wings; black and white wings; gray wings; blue-grey wings and back; long, blue-gray wings; long, pointed wings; long, narrow wings; long, tapered wings; long, dark wings; large wingspan; a forked tail; a long, deeply forked tail; a long tail; a short tail; a long, narrow tail; a long, pointed tail; a short, notched tail; a long, flowing tail; a tail with a distinct V-shape; a black tail; a white tail; a long, black tail; a black tail with white sides; a white tail with a black tip; a black and white barred tail; a tail with white bars; a small bird; a small, plump bird; a medium-sized bird; a large bird; a large, stocky body; a large, stocky bird; a slender body; a long, slender body; a plump body; a stocky body; a round body; a small, compact body; long legs; short legs; webbed feet; large webbed feet; long, webbed feet; orange legs; orange legs and feet; red legs; black legs; black legs and feet; gray legs; yellow legs and feet; pale pink legs; greenish-yellow legs; long, black legs; a long neck; a long, graceful neck; a long, slender neck; a long, curved neck; a white collar around the neck; a black ring around the neck; a black neck and head; red eyes; yellow eyes; orange eyes; blue eyes; a red ring around the eye; a white eye ring; orange or yellow eyes; shiny black eyes; a seabird; a swift, direct flight; a swift, acrobatic flight; shy and secretive behavior; a loud, melodious song; a loud, harsh cry; a raucous call \\
        \bottomrule
    \end{tabular}}
    \caption{\updateremi{\textbf{Classes and supervised concepts for each dataset.} Cats-Dogs-Cars uses 39 concepts, MonumAI 15, COCO 80 (the standard COCO object categories), and CUB-200-2011 226 concepts.}}
    \label{tab:dataset_details}
\end{table*}

\begin{table*}[h]
    \centering
    \small
    \renewcommand{\arraystretch}{1.3}
    \updateremi{\begin{tabular}{llp{0.62\linewidth}}
        \toprule
        \textbf{Dataset} & \textbf{Classes} & \textbf{Concepts} \\
        \midrule
        \multirow{2}{*}{Waterbirds}
        & \multirow{2}{*}{landbird, waterbird}
        & \textit{Non-spurious:} hooked bill, conical bill, thin bill, long bill, curved bill, pointed bill, black cap, colored crown, eye stripe, face mask, crested head, all-black plumage, all-white plumage, iridescent plumage, streaked plumage, spotted plumage, colored breast, white belly, pale underparts, banded breast, wing bars, wing patches, dark wingtips, long wings, pointed wings, forked tail, long tail, short tail, barred tail, small body, large body, stocky body, slender body, long legs, webbed feet, colored legs, long neck, neck collar, streaked back, dark back, colored back, soaring flight, large wingspan, secretive behavior, red eyes, yellow eyes, eye ring, melodious song \\
        \cmidrule(l){3-3}
        & & \textit{Spurious:} green, sea, river, lake, grass, rocks, trees, sky, blue \\
        \midrule
        \multirow{2}{*}{CelebA}
        & \multirow{2}{*}{not blond, blond}
        & \textit{Non-spurious:} blond hair, black hair, brown hair, gray hair, red hair, dark hair, light hair, straight hair, wavy hair, curly hair, long hair, short hair, bangs, bald head, oval face, round face, chubby cheeks, double chin, pale skin, rosy cheeks, high cheekbones, pointy nose, big nose, big lips, arched eyebrows, bushy eyebrows, narrow eyes, bags under eyes, wearing glasses, smiling, mouth open, young face, old face, attractive, receding hairline \\
        \cmidrule(l){3-3}
        & & \textit{Spurious:} feminine face, masculine face, heavy makeup, lipstick, wearing earrings, wearing hat, wearing necklace, wearing necktie, mustache, goatee, beard, sideburns \\
        \bottomrule
    \end{tabular}}
    \caption{\updateremi{\textbf{Classes and supervised concepts for the spurious-correlation benchmarks.} Concepts are split into \textit{non-spurious} (core, target-predictive attributes) and \textit{spurious} (attributes correlated with the spurious factor, namely background for Waterbirds and gender for CelebA). Waterbirds uses 48 non-spurious and 9 spurious concepts; CelebA uses 35 non-spurious and 12 spurious concepts.}}
    \label{tab:spurious_dataset_details}
\end{table*}

\section{\updateremi{Metrics}} \label{appendix:metrics}

\paragraph{Attribution Gini.}

To measure the monosemanticity of nodes, we quantify this with the Gini index of
the per-node attribution vector
$w^{(n)} = \bigl(\mathrm{Shap}^{(n)}(1), \dots, \mathrm{Shap}^{(n)}(M)\bigr)$,
with sorted entries $w^{(n)}_{(1)} \leq \dots \leq w^{(n)}_{(M)}$,
\begin{equation} 
    \mathrm{Gini}(n) \;=\;
    \frac{\sum_{i=1}^{M} (2i - M - 1)\, w^{(n)}_{(i)}}
         {M \sum_{i=1}^{M} w^{(n)}_{(i)}}
    \;\in\; [0, 1].
    \label{eq:attribution-gini}
\end{equation}
A value near $0$ means the node weights all concepts equally, while a value near
$1$ means its attribution concentrates on a single concept. Like coherence, the
index is scale-invariant and thus comparable between our Choquet attributions and
the raw weights of a linear layer; we report it averaged over nodes. The two
metrics are complementary: a node may be sparse yet incoherent, or coherent yet
diffuse, and a desirable node scores high on both.

\paragraph{Node coherence.}
Let $\correlationmatrix \in \mathbb{R}^{M \times M}$ be the correlation
matrix of the concept scores $\simi_j$ over the test set, and let
$w^{(n)} = \bigl(\mathrm{Shap}^{(n)}(1), \dots, \mathrm{Shap}^{(n)}(M)\bigr)$ be
the attribution vector of node $\mathcal{C}^{(n)}$. We define the per-node coherence as
the attribution-weighted average of the pairwise correlations,
\begin{equation}
    \mathrm{Coh}(n) \;=\;
    \frac{\sum_{j \neq l} w^{(n)}_j\, w^{(n)}_l\, \correlationmatrix_{jl}}
         {\sum_{j \neq l} w^{(n)}_j\, w^{(n)}_l}
    \;\in\; [-1, 1],
    \label{eq:coherence_appendix}
\end{equation}
an instance of the similarity-weighted diversity of~\citet{rao1982diversity}.
Weighting each pair by $w^{(n)}_j w^{(n)}_l$ restricts the score to the concepts
the node actually relies on, and the normalization makes it scale-invariant,
hence comparable between our Choquet attributions and the raw weights of a linear
layer. A value near $1$ indicates a node aggregating correlated concepts, near
$0$ a node mixing unrelated ones, and negative a node driven by anti-correlated
concepts. The Node Coherence metric reported in our experiments is the average
$\frac{1}{\mathcal{|N|}}\sum_{n = 1}^N \mathrm{Coh}(n)$ over the set of nodes.

\paragraph{Worst-group accuracy.}
Following~\cite{sagawa2019distributionally}, we partition the test set into groups $g \in \mathcal{G}$ defined by the joint value of the class label $y$ and the spurious attribute $a$ (e.g., bird type and background in Waterbirds). The worst-group accuracy is
\begin{equation}
    \mathrm{Acc}_{\mathrm{wg}} = \min_{g \in \mathcal{G}} \; \frac{1}{|g|} \sum_{i \in g} \mathbf{1}\!\left[\hat{y}_i = y_i\right],
\end{equation}
i.e., the lowest accuracy attained over all groups. In contrast to the average accuracy, which can stay high while a model fails on under-represented groups, $\mathrm{Acc}_{\mathrm{wg}}$ isolates the groups in which the spurious correlation is broken, and therefore directly measures the model's reliance on the spurious feature.

 \newpage
\section{Additional results}

\subsection{Latent space quality} \label{appendix:num_res_latent}

\updateremi{For convenience, we display here the numerical results of Figure~3 in Tables~\ref{tab:results_acc} and~\ref{tab:quality}.}

\begin{table}[H]
  \centering
  \caption{\updateremi{Accuracy (\%) on four classification benchmarks. The best result per dataset is in bold.} Results are averaged over \updateremi{twenty} runs, $\pm$ denoting standard deviations.}
  \label{tab:results_acc}
  \setlength{\tabcolsep}{2pt}
  \updateremi{\begin{tabular}{lcccc}
    \toprule
    \textbf{Architecture} & \textbf{CDC} & \textbf{MonumAI} & \textbf{COCO} & \textbf{CUB-200} \\
    \midrule
    PCBM        & $\mathbf{96.70 \pm 0.72}$  & $58.12 \pm 8.06$           & $53.49 \pm 5.13$           & $18.59 \pm 11.38$ \\
    PSCBM       & $73.96 \pm 3.40$           & $40.10 \pm 9.12$           & $56.93 \pm 0.62$           & $\mathbf{35.84 \pm 12.26}$ \\
    Sparse AE   & $80.19 \pm 4.49$           & $47.16 \pm 4.72$           & $54.46 \pm 0.33$           & $27.32 \pm 1.47$ \\
    SLR-AVD     & $90.37 \pm 17.70$          & $54.57 \pm 10.03$          & $54.28 \pm 4.90$           & $19.41 \pm 10.01$ \\
    CHOQOLATE   & $95.12 \pm 0.27$           & $\mathbf{68.94 \pm 1.18}$  & $\mathbf{64.15 \pm 0.25}$  & $34.20 \pm 1.56$ \\
    \bottomrule
  \end{tabular}}
\end{table}

\begin{table}[H]
  \centering
  \caption{\updateremi{\textbf{Latent space quality metrics across four benchmarks.} The best result per column is in bold.
    $\uparrow$ higher is better. Results averaged among twenty runs, $\pm$ corresponding to standard deviations.
    }}
  \label{tab:quality}
  \setlength{\tabcolsep}{5pt}
  \updateremi{\resizebox{\linewidth}{!}{%
  \begin{tabular}{l *{2}{c} *{2}{c} *{2}{c} *{2}{c}}
    \toprule
    & \multicolumn{2}{c}{\textbf{Cats/Dogs/Cars}}
    & \multicolumn{2}{c}{\textbf{MonumAI}}
    & \multicolumn{2}{c}{\textbf{COCO}}
    & \multicolumn{2}{c}{\textbf{CUB-200}} \\
    \cmidrule(lr){2-3}\cmidrule(lr){4-5}\cmidrule(lr){6-7}\cmidrule(lr){8-9}
    \textbf{Architecture}
      & \makecell{Attr.\\Gini $\uparrow$}
      & \makecell{Node\\Coh. $\uparrow$}
      & \makecell{Attr.\\Gini $\uparrow$}
      & \makecell{Node\\Coh. $\uparrow$}
      & \makecell{Attr.\\Gini $\uparrow$}
      & \makecell{Node\\Coh. $\uparrow$}
      & \makecell{Attr.\\Gini $\uparrow$}
      & \makecell{Node\\Coh. $\uparrow$} \\
    \midrule
    PCBM
      & $0.386 \pm 0.021$ & $0.549 \pm 0.011$
      & $0.349 \pm 0.034$ & $0.478 \pm 0.021$
      & $0.404 \pm 0.086$ & $0.408 \pm 0.005$
      & $0.482 \pm 0.088$ & $0.303 \pm 0.016$ \\
    PSCBM
      & $0.394 \pm 0.034$ & $0.612 \pm 0.059$
      & $0.394 \pm 0.032$ & $\mathbf{0.583 \pm 0.052}$
      & $0.407 \pm 0.010$ & $0.233 \pm 0.010$
      & $0.390 \pm 0.007$ & $0.385 \pm 0.029$ \\
    Sparse AE
      & $0.398 \pm 0.013$ & $0.558 \pm 0.009$
      & $0.345 \pm 0.020$ & $0.511 \pm 0.013$
      & $0.422 \pm 0.019$ & $0.399 \pm 0.004$
      & $0.433 \pm 0.006$ & $0.303 \pm 0.002$ \\
    SLR-AVD
      & $0.394 \pm 0.031$ & $0.550 \pm 0.012$
      & $0.356 \pm 0.031$ & $0.478 \pm 0.025$
      & $0.425 \pm 0.061$ & $0.408 \pm 0.005$
      & $0.321 \pm 0.068$ & $0.289 \pm 0.009$ \\
    CHOQOLATE
      & $\mathbf{0.924 \pm 0.012}$ & $\mathbf{0.680 \pm 0.020}$
      & $\mathbf{0.701 \pm 0.076}$ & $0.487 \pm 0.036$
      & $\mathbf{0.779 \pm 0.017}$ & $\mathbf{0.409 \pm 0.006}$
      & $\mathbf{0.982 \pm 0.006}$ & $\mathbf{0.490 \pm 0.051}$ \\
    \bottomrule
  \end{tabular}}}
\end{table}

\updateremi{In addition, we report in Table~\ref{tab:appendix_cknna_hsic} two additional metrics, CKNNA~\cite{huh2024platonic} and HSIC~\cite{gretton2005measuring}, which both quantify the amount of information preserved between the input embedding (the CLIP-score vector $Z$) and the learned intermediate representation. These metrics are commonly used in representation-learning literature as proxies for how faithfully a layer preserves the structure of its input.}

\updateremi{We deliberately exclude these two quantities from the main-paper evaluation, as they do not measure interpretability or representation quality in the sense we target. Rather, they answer a distinct question: \emph{should the intermediate latent representation maximize information sharing with the input?} As discussed in Section~1, a good concept bottleneck representation must preserve enough information to maintain prediction quality, which corresponds to maximizing mutual information with the \emph{output} distribution, but does not necessarily require maximizing mutual information with the \emph{input}. In fact, an over-faithful preservation of the input is at odds with the goal of compressing redundant concepts into a low-dimensional, semantically coherent representation: a method that simply copies its input would obtain near-perfect CKNNA and HSIC while providing no interpretability benefit at all.}

\updateremi{We therefore report these metrics for transparency rather than as a basis for ranking. As shown in Table~\ref{tab:appendix_cknna_hsic}, methods that operate close to the identity (such as Sparse AE on simple datasets) tend to score highly, while CHOQOLATE trades some of this raw input-preservation for the structural properties (sparsity, semantic coherence) reported in the main paper.}

\begin{table}[H]
    \centering
    \caption{\updateremi{\textbf{CKNNA and HSIC across architectures and datasets.} Both metrics quantify the amount of information shared between the CLIP-score input and the learned latent representation. High values indicate that the representation preserves the input structure but do not, on their own, indicate good interpretability or concept organization (see text).}}
    \label{tab:appendix_cknna_hsic}
    \setlength{\tabcolsep}{5pt}
    \updateremi{\resizebox{\linewidth}{!}{%
    \begin{tabular}{l *{2}{c} *{2}{c} *{2}{c} *{2}{c}}
        \toprule
        & \multicolumn{2}{c}{\textbf{Cats/Dogs/Cars}}
        & \multicolumn{2}{c}{\textbf{MonumAI}}
        & \multicolumn{2}{c}{\textbf{COCO}}
        & \multicolumn{2}{c}{\textbf{CUB-200}} \\
        \cmidrule(lr){2-3}\cmidrule(lr){4-5}\cmidrule(lr){6-7}\cmidrule(lr){8-9}
        \textbf{Architecture}
          & HSIC $\uparrow$ & CKNNA $\uparrow$
          & HSIC $\uparrow$ & CKNNA $\uparrow$
          & HSIC $\uparrow$ & CKNNA $\uparrow$
          & HSIC $\uparrow$ & CKNNA $\uparrow$ \\
        \midrule
        PCBM
          & $0.588 \pm 0.130$ & $0.389 \pm 0.262$
          & $0.540 \pm 0.180$ & $0.331 \pm 0.272$
          & $0.088 \pm 0.051$ & $0.000 \pm 0.000$
          & $0.745 \pm 0.116$ & $0.328 \pm 0.220$ \\
        PSCBM
          & $0.779 \pm 0.009$ & $0.485 \pm 0.112$
          & $0.874 \pm 0.015$ & $0.557 \pm 0.035$
          & $0.499 \pm 0.013$ & $0.142 \pm 0.041$
          & $0.938 \pm 0.011$ & $0.553 \pm 0.132$ \\
        Sparse AE
          & $0.945 \pm 0.008$ & $1.084 \pm 0.173$
          & $0.845 \pm 0.035$ & $0.710 \pm 0.359$
          & $0.769 \pm 0.105$ & $0.082 \pm 0.080$
          & $0.887 \pm 0.010$ & $1.113 \pm 0.149$ \\
        SLR-AVD
          & $0.490 \pm 0.209$ & $0.279 \pm 0.301$
          & $0.460 \pm 0.205$ & $0.204 \pm 0.233$
          & $0.097 \pm 0.040$ & $0.000 \pm 0.000$
          & $0.725 \pm 0.220$ & $0.394 \pm 0.316$ \\
        CHOQOLATE
          & $0.924 \pm 0.035$ & $0.739 \pm 0.151$
          & $0.927 \pm 0.024$ & $0.708 \pm 0.125$
          & $0.950 \pm 0.014$ & $0.666 \pm 0.139$
          & $0.706 \pm 0.047$ & $0.130 \pm 0.074$ \\
        \bottomrule
    \end{tabular}}}
\end{table}
\subsection{\updateremi{Ablation study}} \label{appendix:ablation}

\paragraph{\updateremi{Cross-entropy temperature}}

\begin{figure}[ht]
  \begin{minipage}[c]{0.52\linewidth}
    \centering
    \captionof{table}{\updateremi{\textbf{Effect of temperature $T_{\mathrm{ce}}$ on CHOQOLATE on COCO.} Accuracy (\%) and latent space quality metrics reported on COCO. Results are averaged over five runs, $\pm$ denoting standard deviations.}}
    \label{tab:ablation_temp}
    \setlength{\tabcolsep}{4pt}
    \small
    \updateremi{%
    \begin{tabular}{c ccc}
        \toprule
        $T_{\mathrm{ce}}$
        & \makecell{Acc. $\uparrow$}
        & \makecell{Attr.\\Gini $\uparrow$}
        & \makecell{Node\\Coh. $\uparrow$} \\
        \midrule
        $0.001$ & $65.54 \pm 0.24$ & $0.377 \pm 0.016$ & $0.402 \pm 0.002$ \\
        $0.002$ & $65.37 \pm 0.16$ & $0.482 \pm 0.025$ & $0.403 \pm 0.002$ \\
        $0.005$ & $64.93 \pm 0.14$ & $0.694 \pm 0.008$ & $0.397 \pm 0.003$ \\
        $0.01$  & $63.60 \pm 0.41$ & $0.828 \pm 0.011$ & $0.410 \pm 0.008$ \\
        $0.05$  & $51.93 \pm 0.80$ & $0.968 \pm 0.003$ & $0.457 \pm 0.029$ \\
        $0.1$   & $51.05 \pm 1.19$ & $0.973 \pm 0.006$ & $0.462 \pm 0.030$ \\
        $1.0$   & $49.38 \pm 0.62$ & $0.955 \pm 0.013$ & $0.435 \pm 0.018$ \\
        \bottomrule
    \end{tabular}%
    }
  \end{minipage}
  \hfill
  \begin{minipage}[c]{0.35\linewidth}
    \centering
    \includegraphics[width=0.6\linewidth,keepaspectratio]{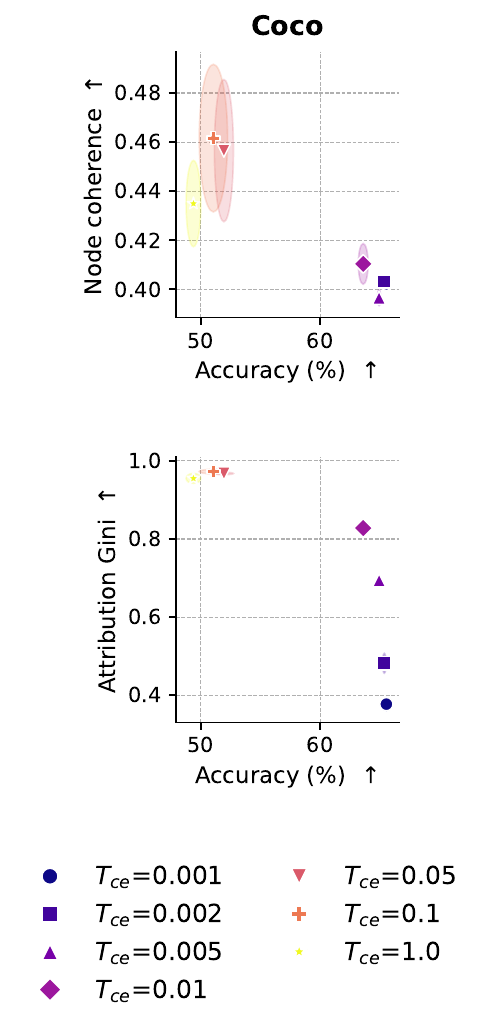}
    \captionof{figure}{\updateremi{\textbf{Accuracy vs.\ organisation quality metrics for different values of $T_{\mathrm{ce}}$.} Each row corresponds to a quality metric. Each point represents a method averaged over five runs, ellipses representing variance bounds.}}
    \label{fig:ablation_temperature}
  \end{minipage}
\end{figure}

\updateremi{We propose here to observe the implication of the hyperparameter $T_{ce}$ by drawing accuracy / latent space quality plots for several values: $0.001,0.002,0.005,0.01,0.05,0.1$ and $1$. Results are available on Table \ref{tab:ablation_temp} and Figure~\ref{fig:ablation_temperature}. Experiments are performed for the standard 2-additive Choquet integral, without $\ell_1$ regularisation.}

\updateremi{Across nearly all configurations, $T_{\text{ce}}$ consistently controls the interpretability--accuracy trade-off. Consistent with the analysis of Section~5, large values of $T_{\text{ce}}$ drive the network toward overly diffuse, low-confidence predictions that sustain sparsity-promoting gradients at the cost of accuracy, while small values cause the error signal to collapse prematurely, yielding dense, non-sparse weights. Notably, setting $T_{\text{ce}} = 1$, which amounts to removing this hyperparameter entirely, falls into the latter regime, confirming the necessity of its introduction.}

\paragraph{\updateremi{Number of classes on CUB-200}}

\begin{figure}[ht]
  \begin{minipage}[c]{0.52\linewidth}
    \centering
    \captionof{table}{\updateremi{\textbf{Effect of the number of classes on CHOQOLATE on CUB-200.} Accuracy (\%) and latent space quality metrics reported on CUB-200 subsets of increasing size. Results are averaged over five runs, $\pm$ denoting standard deviations.}}
    \label{tab:ablation_nclasses}
    \setlength{\tabcolsep}{4pt}
    \small
    \updateremi{%
    \begin{tabular}{c ccc}
        \toprule
        \makecell{\# classes}
        & \makecell{Acc. $\uparrow$}
        & \makecell{Attr.\\Gini $\uparrow$}
        & \makecell{Node\\Coh. $\uparrow$} \\
        \midrule
        $5$   & $86.52 \pm 2.46$ & $0.365 \pm 0.158$ & $0.430 \pm 0.007$ \\
        $10$  & $88.40 \pm 1.92$ & $0.467 \pm 0.072$ & $0.397 \pm 0.003$ \\
        $20$  & $88.47 \pm 1.28$ & $0.732 \pm 0.017$ & $0.354 \pm 0.009$ \\
        $50$  & $76.05 \pm 0.94$ & $0.844 \pm 0.016$ & $0.307 \pm 0.010$ \\
        $100$ & $67.58 \pm 1.00$ & $0.876 \pm 0.015$ & $0.318 \pm 0.009$ \\
        $200$ & $54.56 \pm 0.86$ & $0.910 \pm 0.012$ & $0.322 \pm 0.003$ \\
        \bottomrule
    \end{tabular}%
    }
  \end{minipage}
  \hfill
  \begin{minipage}[c]{0.35\linewidth}
    \centering
    \includegraphics[width=0.6\linewidth,keepaspectratio]{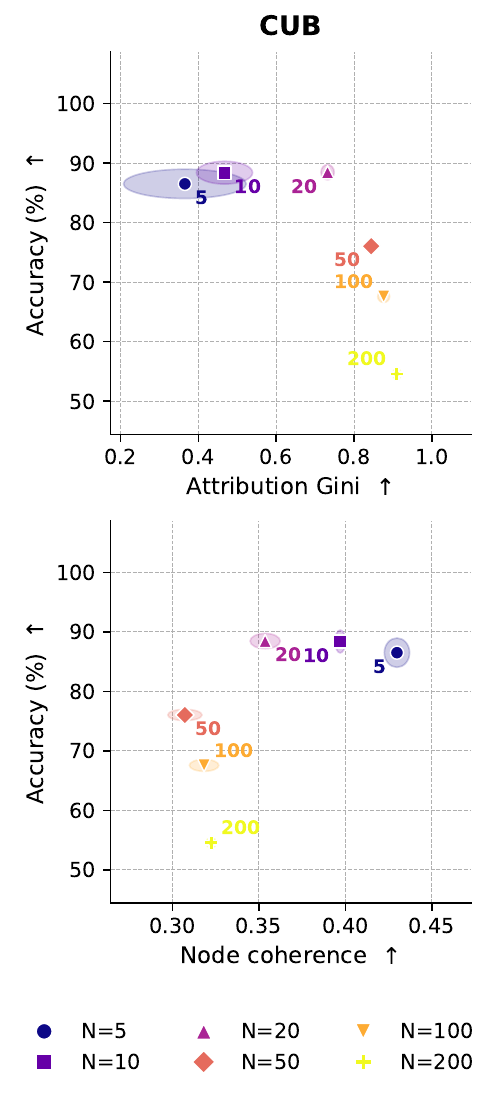}
    \captionof{figure}{\updateremi{\textbf{Accuracy vs.\ organisation quality metrics for different numbers of classes on CUB-200.} Each row corresponds to a quality metric. Each point represents a method averaged over five runs, ellipses representing variance bounds.}}
    \label{fig:ablation_nclasses}
  \end{minipage}
\end{figure}

\updateremi{We investigate here how the difficulty of the classification task, controlled by the number of classes, affects the accuracy / latent space quality trade-off. To this end, we train CHOQOLATE on subsets of CUB-200~\citep{WahCUB_200_2011} comprising $5, 10, 20, 50, 100$ and $200$ classes. Results are available in Table~\ref{tab:ablation_nclasses} and Figure~\ref{fig:ablation_nclasses}. Experiments are performed with the standard configuration adopted in the main paper.}

\updateremi{We observe a clear tension between task difficulty and representation quality. As the number of classes grows, the Attribution Gini increases steadily, indicating that the model concentrates each node on fewer concepts, an effect amplified by the low-dimensional bottleneck having to discriminate among more classes. Accuracy, however, degrades sharply beyond $20$ classes, dropping from $88.47\%$ at $20$ classes to $54.56\%$ at the full $200$ classes, reflecting the limited capacity of the compact representation to accommodate a large label space. Node Coherence follows the opposite trend to Gini, decreasing as more classes are added, which suggests that the concept groups become more heterogeneous as the model is forced to encode finer-grained distinctions. Overall, this experiment delineates the regime in which CHOQOLATE remains effective: a moderate number of classes relative to the bottleneck dimensionality, consistent with the limitations discussed in the main paper.}

\paragraph{\updateremi{CHOQOLATE components}}

\updateremi{We assess the contribution of each component of CHOQOLATE through an ablation study. The components considered are:}

\updateremi{\begin{itemize}
    \item \textbf{1\textsuperscript{st} order}: inclusion of the linear term $\sum_j a_j u_j$ in the Choquet integral.
    \item \textbf{2\textsuperscript{nd} order}: inclusion of the pairwise interaction terms $b_{j,l}\min(u_j, u_l) + c_{j,l}\max(u_j, u_l)$ encoding complementarity and substitutability.
    \item \textbf{Softmax weights}: enforcing the simplex constraint ($a_j,\, b_{j,l},\, c_{j,l} \geq 0$ and $\sum_j a_j + \sum_{j<l}(b_{j,l} + c_{j,l}) = 1$) via a softmax reparametrization.
    \item \textbf{$\ell_1$ inter}: $\ell_1$ regularization applied to the first (intermediate) Choquet layer only.
    \item \textbf{$\ell_1$ all}: $\ell_1$ regularization applied to all Choquet layers.
\end{itemize}}

\begin{figure}[ht]
  \begin{minipage}[c]{0.52\linewidth}
    \centering
    \captionof{table}{\updateremi{\textbf{Ablation study of CHOQOLATE components on COCO.} Each row corresponds to a configuration of the model; checkmarks indicate enabled components. Accuracy (\%) is reported on COCO. Results are averaged over five runs, $\pm$ denoting standard deviations.}}
    \label{tab:ablation_table}
    \setlength{\tabcolsep}{4pt}
    \small
    \updateremi{%
    \begin{tabular}{ccccc ccc}
        \toprule
        \makecell{1\textsuperscript{st}\\order}
        & \makecell{2\textsuperscript{nd}\\order}
        & \makecell{Softmax\\weights}
        & \makecell{$\ell_1$\\inter}
        & \makecell{$\ell_1$\\all}
        & \makecell{Acc. $\uparrow$}
        & \makecell{Attr.\\Gini $\uparrow$}
        & \makecell{Node\\Coh. $\uparrow$} \\
        \midrule
        \checkmark &            &            &            &            
          & $63.11 \pm 0.38$ & $0.416 \pm 0.009$ & $0.397 \pm 0.005$ \\
        \checkmark &            &            &            & \checkmark 
          & $63.12 \pm 0.45$ & $0.418 \pm 0.005$ & $0.395 \pm 0.001$ \\
        \checkmark &            & \checkmark &            &            
          & $64.85 \pm 0.22$ & $0.691 \pm 0.018$ & $0.402 \pm 0.003$ \\
        \checkmark & \checkmark &            &            &            
          & $56.90 \pm 1.60$ & $0.425 \pm 0.013$ & $0.393 \pm 0.003$ \\
        \checkmark & \checkmark & \checkmark &            &            
          & $64.93 \pm 0.14$ & $0.694 \pm 0.008$ & $0.397 \pm 0.003$ \\
        \checkmark & \checkmark & \checkmark & \checkmark &            
          & $64.43 \pm 0.29$ & $0.767 \pm 0.009$ & $0.411 \pm 0.003$ \\
        \checkmark & \checkmark & \checkmark &            & \checkmark 
          & $42.91 \pm 3.44$ & $0.001 \pm 0.000$ & $0.415 \pm 0.000$ \\
        \checkmark & \checkmark &            & \checkmark &            
          & $55.56 \pm 2.05$ & $0.450 \pm 0.021$ & $0.389 \pm 0.004$ \\
        \bottomrule
    \end{tabular}%
    }
  \end{minipage}
  \hfill
  \begin{minipage}[c]{0.35\linewidth}
    \centering
    \includegraphics[width=\linewidth,keepaspectratio]{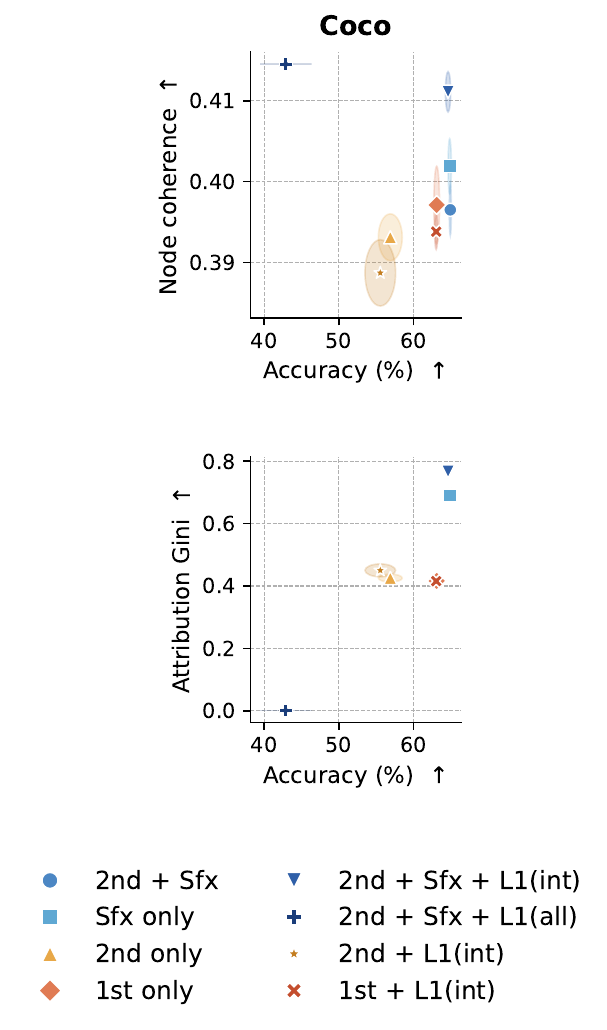}
    \captionof{figure}{\updateremi{\textbf{Accuracy
    vs.\ organisation quality metrics for different variants of CHOQOLATE.} Each row corresponds to a quality metric. Each point represents a method averaged over five runs, ellipses representing variance bounds.}}
    \label{fig:ablation_plot}
  \end{minipage}
\end{figure}

\updateremi{To do so, we trained on multiple variants on COCO. Results are available on Table~\ref{tab:ablation_table}. The visualisation on the accuracy / latent organisation plot is also presented, in Figure~\ref{fig:ablation_plot}}

\updateremi{Overall, row~6 (all components, with $\ell_1$ regularization restricted to the interaction weights of the intermediate layer) achieves the most balanced trade-off on COCO and is therefore adopted as the default configuration throughout the main paper. Notably, the $\ell_1$ penalty and the second-order weights appear to act in synergy: adding the second-order terms to the normalized variant (row~3 vs.\ row~5) leaves both accuracy and latent-space quality essentially unchanged, whereas combining them with the $\ell_1$ penalty raises interpretability at comparable accuracy.}

\paragraph{\updateremi{Number of latent nodes}} \label{appendix:sensitivity_nlatent}

\begin{figure}[ht]
    \centering
    \includegraphics[width=0.62\linewidth,keepaspectratio]{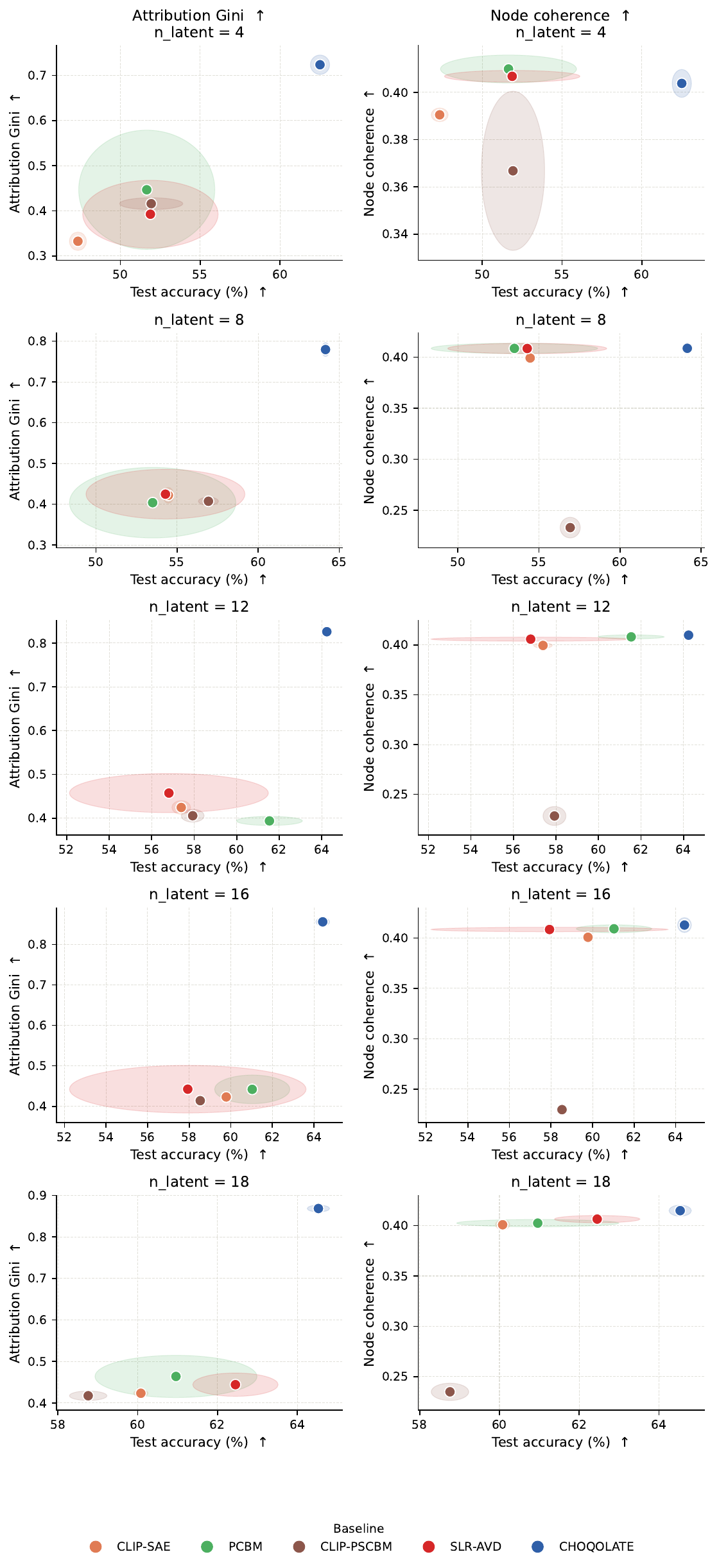}
    \caption{\updateremi{\textbf{Effect of the number of latent nodes $N$ on CHOQOLATE on COCO.} Accuracy vs.\ Attribution Gini (left) and Node Coherence (right), for $N \in \{4, 8, 12, 16, 18\}$ (one value per row). Each point is a method averaged over five runs, with ellipses indicating variance bounds.}}
    \label{fig:ablation_nlatent}
\end{figure}

\updateremi{We study how the size of the interpretable bottleneck, controlled by the number of latent nodes $N$, affects the accuracy / latent-space quality trade-off. We train CHOQOLATE and all baselines on COCO for $N \in \{4, 8, 12, 16, 18\}$, keeping every other hyperparameter fixed to the main-paper configuration. Results are reported in Figure~\ref{fig:ablation_nlatent}, with each row corresponding to a value of $N$ and each column to a quality metric plotted against test accuracy.}

\updateremi{Two regimes emerge. For $N = 4$, the bottleneck is too narrow to accommodate the COCO label space: accuracy drops sharply for every method, and the variance of the quality metrics grows, most visibly for CLIP-PSCBM, whose Node Coherence becomes highly unstable. Beyond this point, from $N = 8$ onward, we tend to observe a stabilisation: CHOQOLATE reaches its accuracy plateau immediately, whereas the baselines need a much wider bottleneck to approach it. CHOQOLATE dominates the Attribution Gini axis at every value of $N$, by a wide margin over all baselines, while simultaneously attaining the highest accuracy. On Node Coherence the margins are tighter: CHOQOLATE remains among the best throughout, whereas CLIP-PSCBM collapses once $N \geq 8$, indicating that its nodes aggregate increasingly unrelated concepts as the bottleneck widens.}

\updateremi{Importantly, increasing $N$ beyond $8$ yields no further accuracy gain for CHOQOLATE and only a marginal improvement in latent-space quality, while making the representation less compact and the word-cloud explanations more redundant across nodes. We therefore fix $N = 8$ in the main paper as the smallest bottleneck that already reaches the accuracy plateau while preserving strong sparsity and coherence, and keeping the explanations relatively simple.}

\subsection{\updateremi{Additional Global Explanations}}
\label{appendix:global_explanations}

\updateremi{
We complement the global explanations presented in Section~6 (Figures~4a and~4b) with analogous visualizations on the two remaining benchmarks, MonumAI and COCO. As in the main paper, each node is represented by a word cloud in which the size of a concept reflects its Shapley contribution to that node. We compare CHOQOLATE against the PCBM on each dataset.
}

\begin{figure}[ht]
    \centering
    \includegraphics[width=\linewidth]{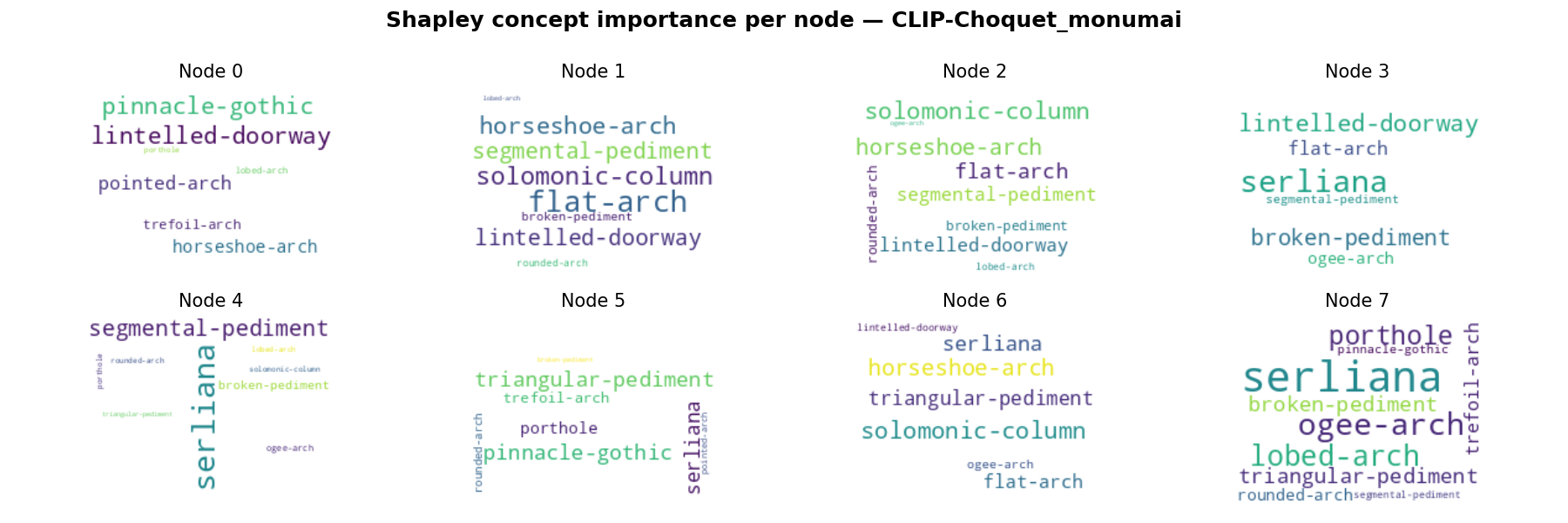}
    \caption{\updateremi{\textbf{Global explanations for PCBM.} Dataset: MonumAI. The size of each word reflects the importance of the corresponding concept for the node.}}
    \label{fig:globalexp_mlp_monumai}
\end{figure}

\begin{figure}[ht]
    \centering
    \includegraphics[width=\linewidth]{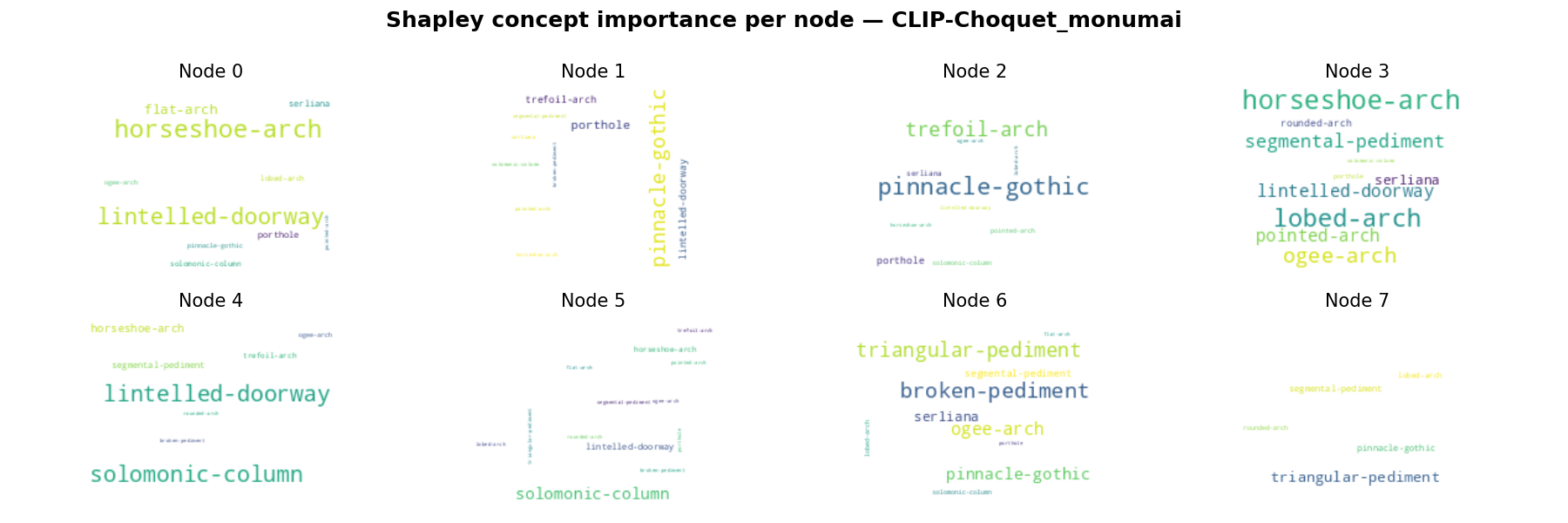}
    \caption{\updateremi{\textbf{Global explanations for CHOQOLATE.} Dataset: MonumAI. The size of each word reflects the importance of the corresponding concept for the node.}}
    \label{fig:globalexp_choqolate_monumai}
\end{figure}

\begin{figure}[ht]
    \centering
    \includegraphics[width=\linewidth]{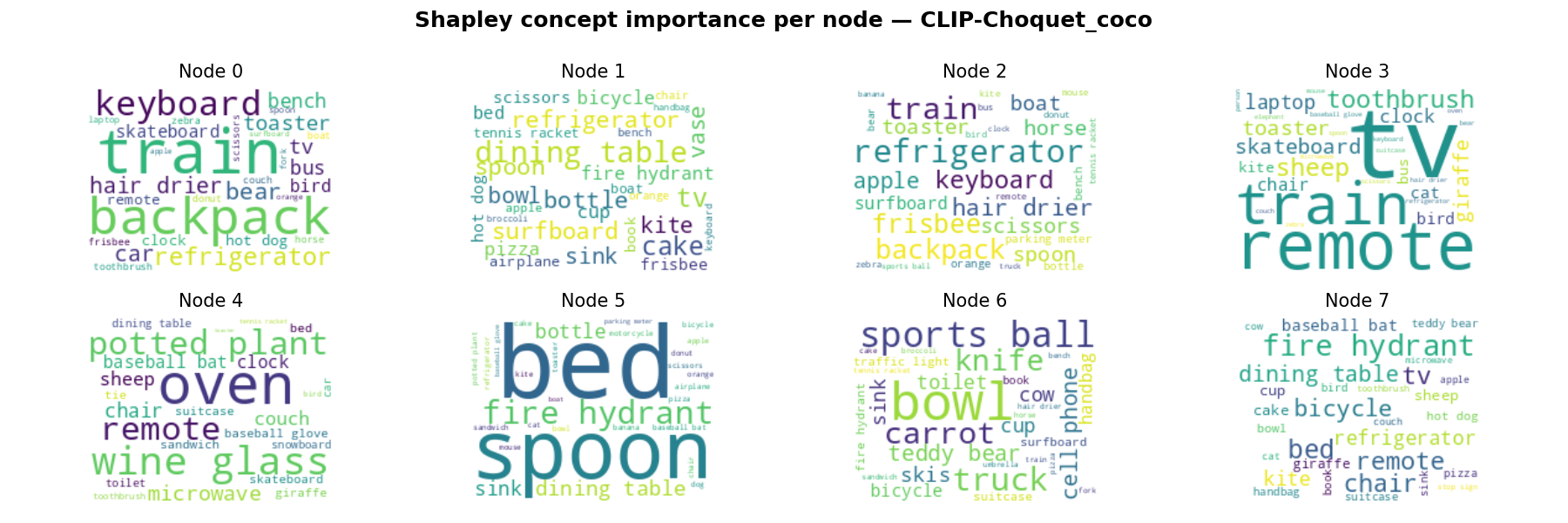}
    \caption{\updateremi{\textbf{Global explanations for the PCBM.} Dataset: COCO. The size of each word reflects the importance of the corresponding concept for the node.}}
    \label{fig:globalexp_mlp_coco}
\end{figure}

\begin{figure}[ht]
    \centering
    \includegraphics[width=\linewidth]{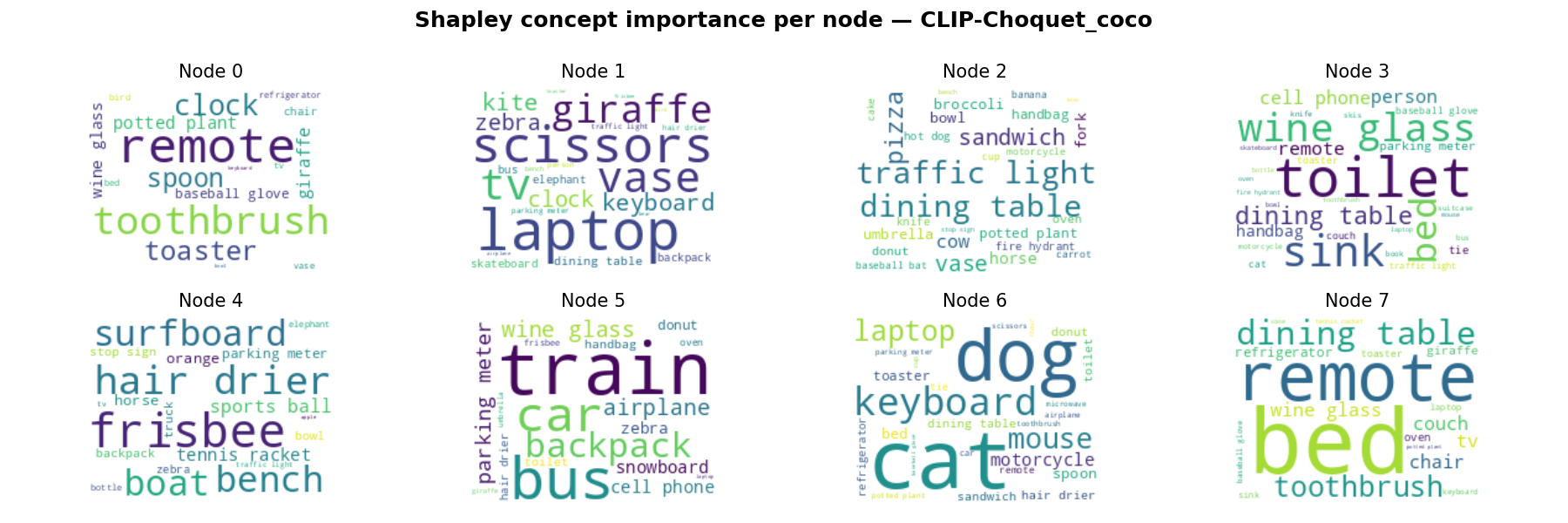}
    \caption{\updateremi{\textbf{Global explanations for CHOQOLATE.} Dataset: COCO. The size of each word reflects the importance of the corresponding concept for the node.}}
    \label{fig:globalexp_choqolate_coco}
\end{figure}

\updateremi{The observations remain globally the same as for Cats/Dogs/Cars. The larger concept set leads to slightly noisier word clouds for both methods, consistent with the more nuanced activation-sparsity behaviour reported in Section~6; nonetheless, CHOQOLATE still produces visibly more salient and semantically grouped explanations than PCBM. For example, in Figure \ref{fig:globalexp_choqolate_coco}, considering that the final objective is to classify types of places, it is interesting to note that the method automatically clustered in node 4 concepts associated with outdoor sports (surfboard, frisbee, boat ...), or transportation in node 5 (train, car, bus ...). }
\newpage

\section{\updateremi{Statistical Significance of the Main Results}}
\label{appendix:significance}

To confirm that the performance differences between CHOQOLATE and each baseline are statistically significant rather than caused by run-to-run variability \updateremi{(Figure~3)}, we perform a one-sided Mann–Whitney U test for every combination of dataset, metric, and baseline.

For each comparison, we test the null hypothesis $H_0$
that the distributions of metric values for CHOQOLATE and the baseline are identical, against the one-sided alternative hypothesis $H_1$ that CHOQOLATE achieves higher metric values than the baseline. A small $p$ value implies rejection of $H_0$, and acceptance of the alternative.

We report the relative median distance $\Delta=(median_{CHOQOLATE}-median_{baseline})/median_{baseline}$ and the p-values in Table~\ref{tab:significance_old} just below. Larger $\Delta$ and smaller $p$ value indicate a more clearly separated, more favorable trade-off for CHOQOLATE:

\begin{table}[H]
   \centering
   \caption{\textbf{Mann Whitney U-test significance of the CHOQOLATE vs.\ baseline separation.} }
    \label{tab:significance_old}
    \setlength{\tabcolsep}{5pt}
    \small
    \updateremi{\begin{tabular}{ll cc cc}
        \toprule
        & & \multicolumn{2}{c}{\textbf{Attr.\ Gini}} 
            & \multicolumn{2}{c}{\textbf{Node Coh.}} \\
        \cmidrule(lr){3-4}\cmidrule(lr){5-6}
        \textbf{Dataset} & \textbf{Baseline}
          & $\Delta$ & $p$
          & $\Delta$ & $p$ \\
        \midrule
        \multirow{4}{*}{Cats/Dogs/Cars}
          & PCBM       & $1.37$ & $<0.001$ & $0.23$  & $<0.001$ \\
          & PSCBM      & $1.35$ & $<0.001$ & $0.12$  & $<0.001$ \\
          & Sparse AE  & $1.33$ & $<0.001$ & $0.21$  & $<0.001$ \\
          & SLR-AVD    & $1.35$ & $<0.001$ & $0.23$  & $<0.001$ \\
        \midrule
        \multirow{4}{*}{MonumAI}
          & PCBM       & $0.99$ & $<0.001$ & $0.01$  & $0.231$ \\
          & PSCBM      & $0.83$ & $<0.001$ & $-0.13$ & $1.000$ \\
          & Sparse AE  & $1.10$ & $<0.001$ & $-0.05$ & $0.994$ \\
          & SLR-AVD    & $1.04$ & $<0.001$ & $0.02$  & $0.157$ \\
        \midrule
        \multirow{4}{*}{COCO}
          & PCBM       & $1.06$ & $<0.001$ & $0.01$ & $0.579$ \\
          & PSCBM      & $0.91$ & $<0.001$ & $0.76$  & $<0.001$ \\
          & Sparse AE  & $0.85$ & $<0.001$ & $0.02$  & $<0.001$ \\
          & SLR-AVD    & $0.85$ & $<0.001$ & $0.01$ & $0.505$ \\
        \midrule
        \multirow{4}{*}{CUB-200}
          & PCBM       & $1.07$ & $<0.001$ & $0.64$  & $<0.001$ \\
          & PSCBM      & $1.53$ & $<0.001$ & $0.29$  & $<0.001$ \\
          & Sparse AE  & $1.27$ & $<0.001$ & $0.63$  & $<0.001$ \\
          & SLR-AVD    & $2.13$ & $<0.001$ & $0.73$  & $<0.001$ \\
        \bottomrule
    \end{tabular}}
\end{table}

\updateremi{On Attribution Gini, the separation between CHOQOLATE and every
baseline is statistically significant across all four datasets ($p < 0.001$),
with consistently large positive median distances ($\Delta$ between $0.83$ and
$2.13$). This confirms that CHOQOLATE's sparsity advantage is robust and not
attributable to run-to-run variability. CHOQOLATE also improves Node Coherence
over the baselines, significantly so on Cats/Dogs/Cars and CUB-200 and, on COCO,
by a wide margin over PSCBM. The median distances on this metric are smaller than
for Attribution Gini, and on MonumAI the methods are statistically comparable,
consistent with the main-paper finding that CHOQOLATE's coherence advantage is
more modest than its sparsity advantage and narrows on datasets whose concept
vocabularies are harder to cluster.}
\newpage

\section{Proof for a Single Choquet Layer} \label{appendix:gradient_derivation}

Throughout this section, $\vecU = (u_1, \dots, u_p) \in [0,1]^p$ denotes the
input of a Choquet layer. Each class $\classeName \in \{1, \dots, C\}$ is assigned a 2-additive Choquet
integral $\CI^{(\classeName)}$ (see Eq.~(2)) whose weights
$$w^{(\classeName)}
\;=\;
\bigl(
\underbrace{a_1^{(\classeName)}, \dots, a_p^{(\classeName)}}_{\text{first-order terms}}\;,
\underbrace{b_{1,2}^{(\classeName)}, \dots, b_{p-1,p}^{(\classeName)}}_{\text{min interactions}}\;,
\underbrace{c_{1,2}^{(\classeName)}, \dots, c_{p-1,p}^{(\classeName)}}_{\text{max interactions}}
\bigr)$$

The 2-additive Choquet integral can be view as a linear function of a feature vector constructed from the input. We build the vector $\phi(\vecU)$, which contains the $p$ coordinates themselves, followed by all the $\min(u_j, u_l)$ and all the $\max(u_j, u_l)$ terms associated with pairs of coordinates.  Therefore, Eq.~(2) can be written as the inner product between this feature vector and the class weight vector:
\begin{align*}
\CI^{(\classeName)}(\vecU) &\;=\; \sum_{j=1}^{p} a_j, u_j
\;+\; \sum_{j < l} \Bigl( b_{j,l}\, \min(u_j, u_l)
\;+\;c_{j,l}\, \max(u_j, u_l) \Bigr) \\
& \;=\; \langle w^{(\classeName)}, \phi(\vecU) \rangle  
\end{align*}
where
$
\phi(\vecU)
\;=\;
\bigl(
\underbrace{u_1, \dots, u_p}_{p \text{ coordinates}}\;,
\underbrace{\min(u_1, u_2), \dots, \min(u_{p-1}, u_p)}_{\binom{p}{2} \text{ minima}}\;,
\underbrace{\max(u_1, u_2), \dots, \max(u_{p-1}, u_p)}_{\binom{p}{2} \text{ maxima}}
\bigr)$.   
We denote by $q = p + 2\binom{p}{2} = p^2$ the number of components of $\phi(\vecU)$.

\subsection{Proof of Proposition~1 of the Main Paper}

The logit of
class $\classeName$ is $\hh_\classeName = \mathcal{C}^{(\classeName)}(\vecU)$, and
$\hat{y}_\classeName =\frac{\exp(\hh_\classeName)}{\sum_{\classeName'=1}^{\nclasses} \exp(\hh_{\classeName'})}$ is the predicted probability of
class $\classeName$. The target of the sample is a one-hot vector
$\vecY \in \{0,1\}^{\nclasses}$, with coordinates $y_\classeName$ and nonzero coordinate at index
$y_\star \in \{1, \dots, \nclasses\}$. Finally, $\loss$ denotes the
cross-entropy loss; the $\ell_1$ penalty of the training
objective~(4) does not depend on the logits and is omitted from
the gradient computations below. Since the target is one-hot, the logit of the ground-truth class is
$\sum_{r=1}^{C} y_\classeName\, \hh_\classeName$, and the loss reads
\begin{equation}
    \loss \;=\; -\sum_{\classeName=1}^{\nclasses} y_\classeName\, \hh_\classeName
    \;+\; \log \sum_{\classeName=1}^{\nclasses} \exp(\hh_\classeName).
    \label{eq:deriv_loss}
\end{equation}

\emph{Step 1: gradient with respect to the logits.}
We differentiate the two terms of~\eqref{eq:deriv_loss} with respect to a
logit $\hh_\classeName$. In the first term, only the summand of index $\classeName$ depends on
$\hh_\classeName$, contributing $-y_\classeName$. For the second term, the chain rule applied
to the log-sum-exp gives
\begin{equation}
    \frac{\partial}{\partial \hh_\classeName}
    \log \sum_{\classeName'=1}^{\nclasses} \exp(\hh_{\classeName'})
    \;=\; \frac{\exp(\hh_\classeName)}{\sum_{\classeName'=1}^{\nclasses} \exp(\hh_{\classeName'})}
    \;=\; \hat{y}_\classeName.
    \label{eq:deriv_step1_lse}
\end{equation}
Summing the two contributions,
\begin{equation}
    \frac{\partial \loss}{\partial \hh_\classeName}
    \;=\; \hat{y}_\classeName - y_\classeName.
    \label{eq:deriv_step1}
\end{equation}

\emph{Step 2: gradient with respect to the Choquet weights.}
The logit of class $\classeName$ expands as
\begin{align}
    \hh_\classeName
    \;&=\; \langle w^{(\classeName)}, \phi(\vecU) \rangle
    \;=\; \sum_{j=1}^{q} w_j^{(\classeName)}\, \phi_j(\vecU)
    \label{eq:deriv_logit_expand} %
\end{align}
The features
$\phi_j(\vecU)$ are functions of the input only: they involve no weight, so
they are constants with respect to $w^{(\classeName)}$. The logit is therefore linear
in the weights, and differentiating the~Eq.~\eqref{eq:deriv_logit_expand}
with respect to a single weight $w_i^{(\classeName)}$ keeps only the summand of index
$j = i$,
\begin{equation}
    \frac{\partial \hh_\classeName}{\partial w_i^{(\classeName)}}
    \;=\; \phi_i(\vecU).
    \label{eq:deriv_logit_weight}
\end{equation}
Moreover, the weight vector $w^{(\classeName)}$ enters the loss only through the
logit $\hh_\classeName$: the other logits $\hh_{r'}$, $r' \neq r$, are computed from
their own weight vectors $w^{(r')}$. The chain rule from $\loss$ to
$w_i^{(\classeName)}$ therefore goes through the single intermediate variable
$\hh_\classeName$, and combining~\eqref{eq:deriv_step1}
with~\eqref{eq:deriv_logit_weight},
\begin{equation}
    \frac{\partial \loss}{\partial w_i^{(\classeName)}}
    \;=\; \frac{\partial \loss}{\partial \hh_\classeName}\,
          \frac{\partial \hh_\classeName}{\partial w_i^{(\classeName)}}
    \;=\; \bigl( \hat{y}_\classeName - y_\classeName \bigr)\, \phi_i(\vecU).
    \label{eq:deriv_step2}
\end{equation}

\emph{Step 3: gradient with respect to the unconstrained parameters.}
The weights of class $\classeName$ are obtained from the unconstrained parameters by
the softmax map,
\begin{equation}
    w_j^{(\classeName)}
    \;=\; \frac{\exp(\theta_j^{(\classeName)})}{\sum_{m=1}^{q} \exp(\theta_m^{(\classeName)})}
    \;=\; \frac{\exp(\theta_j^{(\classeName)})}{\Omega},
    \qquad
    \Omega \;:=\; \sum_{m=1}^{q} \exp(\theta_m^{(\classeName)}),
    \label{eq:deriv_softmax_def}
\end{equation}
so each weight $w_j^{(\classeName)}$ depends on \emph{every} coordinate of
$\theta^{(\classeName)}$ through the normalizer $\Omega$. Differentiating the
quotient~\eqref{eq:deriv_softmax_def} with respect to $\theta_i^{(\classeName)}$, the
numerator contributes $\exp(\theta_j^{(\classeName)})\, \delta_{ji}$ (it depends on
$\theta_i^{(\classeName)}$ only when $j = i$) and the normalizer contributes
$\partial \Omega / \partial \theta_i^{(\classeName)} = \exp(\theta_i^{(\classeName)})$, so
\begin{equation}
    \frac{\partial w_j^{(\classeName)}}{\partial \theta_i^{(\classeName)}}
    \;=\; \frac{\exp(\theta_j^{(\classeName)})\, \delta_{ji}}{\Omega}
    \;-\; \frac{\exp(\theta_j^{(\classeName)})\, \exp(\theta_i^{(\classeName)})}{\Omega^2}
    \;=\; w_j^{(\classeName)} \bigl( \delta_{ji} - w_i^{(\classeName)} \bigr),
    \label{eq:deriv_jacobian}
\end{equation}
where $\delta_{ji}$ is the Kronecker delta and the last equality recognizes
$w_j^{(\classeName)} = \exp(\theta_j^{(\classeName)})/\Omega$ and $w_i^{(\classeName)} = \exp(\theta_i^{(\classeName)})/\Omega$.

Because a single parameter $\theta_i^{(\classeName)}$ moves \emph{all} the weights
$w_1^{(\classeName)}, \dots, w_q^{(\classeName)}$ at once, the chain rule from $\loss$ to
$\theta_i^{(\classeName)}$ sums over the $q$ weights,
\begin{equation}
    \frac{\partial \loss}{\partial \theta_i^{(\classeName)}}
    \;=\; \sum_{j=1}^{q}
          \frac{\partial \loss}{\partial w_j^{(\classeName)}}\,
          \frac{\partial w_j^{(\classeName)}}{\partial \theta_i^{(\classeName)}}
    \;=\; \bigl( \hat{y}_\classeName - y_\classeName \bigr)
          \sum_{j=1}^{q} \phi_j(\vecU)\, w_j^{(\classeName)}
          \bigl( \delta_{ji} - w_i^{(\classeName)} \bigr),
    \label{eq:deriv_cNamehain}
\end{equation}
where the second equality substitutes~\eqref{eq:deriv_step2}
and~\eqref{eq:deriv_jacobian}, and factors $(\hat{y}_\classeName - y_\classeName)$ out of the
sum since it does not depend on $j$.

\emph{Step 4: simplifying the inner sum.}
Splitting the sum along the Kronecker delta,
\begin{equation}
    \sum_{j=1}^{q} \phi_j(\vecU)\, w_j^{(\classeName)} \bigl( \delta_{ji} - w_i^{(\classeName)} \bigr)
    \;=\; w_i^{(\classeName)}\, \phi_i(\vecU)
    \;-\; w_i^{(\classeName)} \sum_{j=1}^{q} w_j^{(\classeName)}\, \phi_j(\vecU)
    \;=\; w_i^{(\classeName)} \bigl( \phi_i(\vecU) - \mathcal{C}^{(\classeName)}(\vecU) \bigr),
    \label{eq:deriv_factored}
\end{equation}
since the first term retains only $j = i$ and the second sum is the Choquet
output.  Substituting this expression into Eq.~\eqref{eq:deriv_cNamehain} yields
\begin{equation*}
    \frac{\partial\loss}{\partial \theta_i^{(\cls)}}
    \;=\; \bigl(\hat{y}_\cls - y_\cls\bigr) \cdot w_i^{(\cls)} \cdot
    \bigl( \phi_i(\vecU) - \CI^{(\cls)}(\vecU) \bigr).
\end{equation*}

This concludes the proof.\hfill$\qed$

\subsection{Proof of Corollary~1}

\begin{corollary}[Effect of temperature scaling] \label{prop:choquet_gradient_temp}
Consider the one-layer Choquet classifier trained with the temperature-scaled
cross-entropy $\loss^{(T_{\mathrm{ce}})}$ for a temperature $T_{\mathrm{ce}} > 0$,
under the hypotheses of Proposition~1. Then
\begin{equation}
    \frac{\partial \loss^{(T_{\mathrm{ce}})}}{\partial \theta_i^{(\cls)}}
    \;=\; \frac{1}{T_{\mathrm{ce}}}
          \bigl( \hat{y}_\cls^{(T_{\mathrm{ce}})} - y_\cls \bigr)\,
          w_i^{(\cls)} \bigl( \phi_i(\vecU) - \mathcal{C}^{(\cls)}(\vecU) \bigr),
    \label{eq:deriv_result_temp}
\end{equation}
where $\hat{y}_\cls^{(T_{\mathrm{ce}})}
    \;=\; \mathrm{softmax}\!\left(\frac{\hh_\cls}{T_{\mathrm{ce}}}\right)$.
\end{corollary}

The prefactor amounts to a change of learning rate; the effect sits in
$\hat{y}_\cls^{(T_{\mathrm{ce}})}$. A small $T_{\mathrm{ce}}$ sharpens
predictions and kills the error term, and the sparsity pressure with it, as
soon as the model gets confident; a large one keeps the signal alive at the
cost of slower convergence. Since Choquet outputs live in $[0,1]$, logit
gaps never exceed one: useful temperatures sit well below one, where
$T_{\mathrm{ce}}$ trades interpretability against accuracy
(Appendix~\ref{appendix:ablation}).

With temperature, the loss (Eq.~\eqref{eq:deriv_loss}) is evaluated on the
rescaled logits,
\begin{equation}
    \loss
    \;=\; -\sum_{r=1}^{\nclasses} y_\classeName\, \frac{\hh_\classeName}{T_{\mathrm{ce}}}
    \;+\; \log \sum_{r=1}^{\nclasses} \exp\!\left(\frac{\hh_\classeName}{T_{\mathrm{ce}}}\right).
    \label{eq:deriv_loss_temp}
\end{equation}
We differentiate with respect to $\hh_\classeName$, using
$\partial (\hh_\classeName / T_{\mathrm{ce}}) / \partial \hh_\classeName = 1 / T_{\mathrm{ce}}$.
In the first term, only the summand of index $\classeName$ depends on $\hh_\classeName$,
contributing $-y_\classeName / T_{\mathrm{ce}}$. For the second term, the chain rule
applied to the log-sum-exp gives, exactly as
in~\eqref{eq:deriv_step1_lse} but with rescaled arguments,
\begin{equation}
    \frac{\partial}{\partial \hh_\classeName}
    \log \sum_{r'=1}^{\nclasses} \exp\!\left(\frac{\hh_{r'}}{T_{\mathrm{ce}}}\right)
    \;=\; \frac{1}{T_{\mathrm{ce}}}\,
    \frac{\exp(\hh_\classeName / T_{\mathrm{ce}})}
         {\sum_{r'=1}^{\nclasses} \exp(\hh_{r'} / T_{\mathrm{ce}})}
    \;=\; \frac{1}{T_{\mathrm{ce}}}\, \hat{y}_\classeName^{(T_{\mathrm{ce}})}.
    \label{eq:deriv_step1_lse_temp}
\end{equation}
Summing the two contributions, Step~1 becomes
\begin{equation}
    \frac{\partial \loss}{\partial \hh_\classeName}
    \;=\; \frac{1}{T_{\mathrm{ce}}}
          \bigl( \hat{y}_\classeName^{(T_{\mathrm{ce}})} - y_\classeName \bigr).
    \label{eq:deriv_step1_temp}
\end{equation}
Steps~2 to~4 concern only the map from $\theta^{(\classeName)}$ to $\hh_\classeName$, namely
$\hh_\classeName = \langle w^{(\classeName)}, \phi(\vecU) \rangle$ with
$w^{(\classeName)} = \mathrm{softmax}(\theta^{(\classeName)})$, in which $T_{\mathrm{ce}}$ plays
no role: the temperature acts on the classification softmax over the
\emph{logits}, not on the reparametrization softmax over the \emph{weights}.
The factor $\frac{1}{T_{\mathrm{ce}}} \bigl( \hat{y}_\classeName^{(T_{\mathrm{ce}})} -
y_\classeName \bigr)$ therefore carries through Steps~2 to~4 in place of
$(\hat{y}_\classeName - y_\classeName)$, yielding~\eqref{eq:deriv_result_temp}. \hfill$\qed$

\section{\updateremi{Technical details}}

\updateremi{All experiments were run on a single NVIDIA GeForce RTX~4070~SUPER GPU (12\,GB of VRAM), using PyTorch on a Linux machine.}

\end{document}